\documentclass{article}

\usepackage{amsmath,amssymb, amsthm}
\usepackage{mathabx}
\usepackage{soul}
\usepackage{pgfplots}
\usepackage{natbib}
\pgfplotsset{width=10cm,compat=1.9}
\usepgfplotslibrary{external}
\usepackage{amsopn}
\DeclareMathOperator*{\argmin}{argmin}
\DeclareMathOperator*{\argmax}{argmax}

\theoremstyle{plain}
\newtheorem{theorem}{Theorem}[section]

\newtheorem{lemma}[theorem]{Lemma}

\theoremstyle{definition}
\newtheorem{definition}[theorem]{Definition}
\newtheorem{assumption}[theorem]{Assumption}
\theoremstyle{remark}

\newcommand{\ZZ}{\mathcal{Z}}

\newcommand{\twonorm}[1]{\left|\left|#1\right|\right|}
\usepackage{hyperref} 
\usepackage{cleveref}
\crefname{algorithm}{Algorithm}{Algorithms}
\crefname{assumption}{Assumption}{Assumptions}
\crefname{equation}{}{}
\crefname{figure}{Fig.}{Figs.}
\crefname{table}{Table}{Tables}
\crefname{section}{Section}{Sections}
\crefname{subsection}{Section}{Sections}
\crefname{theorem}{Theorem}{Theorems}
\crefname{lemma}{Lemma}{Lemmmas}
\crefname{proposition}{Proposition}{Propositions}
\crefname{definition}{Definition}{Definitions}
\crefname{corollary}{Corollary}{Corollaries}
\crefname{remark}{Remark}{Remarks}
\crefname{example}{Example}{Examples}
\crefname{appendix}{Appendix}{Appendices}

\usepackage{amsmath}
\usepackage{amsfonts}
\usepackage{bbm}
\usepackage{color}
\usepackage{graphicx}
\usepackage{enumitem}
\usepackage{wrapfig}
\usepackage{svg}
\usepackage{algorithm}
\usepackage{algorithmic}
\newcommand{\XX}{\mathcal{X}}
\newcommand{\FF}{\mathcal{F}}
\newcommand{\WW}{\mathcal{W}}
\newcommand{\YY}{\mathcal{Y}}

\newcommand{\ws}{w^{*}}

\newcommand{\by}{\mathbf{y}}

\newcommand{\expec}{\mathbb{E}}

\usepackage{mathtools}

\newcommand{\Al}{\mathcal{A}}

\newcommand{\rdt}{\mathbb{R}^{d_\theta}}
\newcommand{\tth}{\theta}
\newcommand{\lw}{L_w}
\newcommand{\lt}{L_\theta}
\newcommand{\bw}{\beta_w}
\newcommand{\bt}{\beta_{\theta}}
\newcommand{\btw}{\beta_{\theta w}}
\newcommand{\kw}{\kappa_w}
\newcommand{\ktw}{\kappa_{\theta w}}
\newcommand{\naw}{\nabla_{w}}
\newcommand{\nt}{\nabla_{\theta}}

\newcommand{\dph}{\Delta_{\Phi}}
\newcommand{\dt}{d_{\tth}}
\newcommand{\dw}{d_w}
\newcommand{\tg}{\widetilde{g}}

\newcommand{\ett}{\eta_{\theta}}
\newcommand{\etw}{\eta_{w}}
\newcommand{\one}{\mathbbm{1}_{\{m < n\}}}
\newcommand{\oneprime}{\mathbbm{1}_{\{m < \tn\}}}
\newcommand{\hy}{\widehat{Y}}

\DeclareMathOperator{\Tr}{Tr}
\newcommand{\bs}{\mathbf{s}}
\newcommand{\haty}{\widehat{y}}

\newcommand{\tn}{\tilde{n}}
\usepackage{arxiv}

\usepackage[utf8]{inputenc} %
\usepackage[T1]{fontenc}    %
\usepackage{hyperref}       %
\usepackage{url}            %
\usepackage{booktabs}       %
\usepackage{nicefrac}       %
\usepackage{microtype}      %
\usepackage{lipsum}		%
\usepackage{graphicx}
\usepackage{doi}
\usepackage{float}
\usepackage{graphicx}
\usepackage{subfig}
\usepackage{placeins}

\title{A Stochastic Optimization Framework for Private and Fair Learning From Decentralized Data}

\date{} 					%

\author{{\hspace{1mm}Devansh Gupta} \\
	Department of Computer Science\\
	University of Southern California\\
	Los Angeles, CA 90007\\
	\texttt{guptadev@usc.edu} \\
	\And
	{\hspace{1mm}A.S. Poornash} \thanks{Work done as a Visiting Scholar at University of Southern California.} \\
	Department of Computer Science\\
	Indian Institute of Technology Patna\\
	Patna, Bihar 801106\\
	\texttt{poornash\_2101cs01@iitp.ac.in} \\
        \And
	{\hspace{1mm}Andrew Lowy} \\
	Department of Computer Science\\
	University of Wisconsin-Madison\\
	Madison, WI 53706\\
	\texttt{alowy@wisc.edu} \\
        \And
	{\hspace{1mm}Meisam Razaviyayn} \\
	Departments of ISE, CS, ECE, and QCB\\
	University of Southern California\\
	Los Angeles, CA 90007\\
	\texttt{razaviya@usc.edu} \\
}

\renewcommand{\shorttitle}{Stochastic Private and Fair Learning From Decentralized Data}

\hypersetup{
pdftitle={A template for the arxiv style},
pdfsubject={q-bio.NC, q-bio.QM},
pdfauthor={David S.~Hippocampus, Elias D.~Striatum},
pdfkeywords={First keyword, Second keyword, More},
}

\begin{document}
\maketitle

\begin{abstract}
	Machine learning models are often trained on sensitive data (e.g., medical records and race/gender) that is distributed across different ``silos'' (e.g., hospitals). These \textit{federated learning} models may then be used to make consequential decisions, such as allocating healthcare resources. Two key challenges emerge in this setting: (i) maintaining the \textit{privacy} of each person's data, even if other silos or an adversary with access to the central server tries to infer this data; (ii) ensuring that decisions are \textit{fair} to different demographic groups (e.g., race/gender). In this paper, we develop a novel algorithm for private and fair federated learning (FL). Our algorithm satisfies \textit{inter-silo record-level differential privacy} (ISRL-DP), a strong notion of private FL requiring that silo $i$'s sent messages satisfy record-level differential privacy for all $i$. Our framework can be used to promote different fairness notions, including demographic parity and equalized odds. We prove that our algorithm converges under mild smoothness assumptions on the loss function, whereas prior work required strong convexity for convergence. 
As a byproduct of our analysis, we obtain the first convergence guarantee for ISRL-DP nonconvex-strongly concave min-max FL. Experiments 
demonstrate the state-of-the-art fairness-accuracy tradeoffs of our algorithm across different privacy levels. 
\end{abstract}


\section{Introduction}
\label{sec: intro}
Many important decisions are being assisted by machine learning (ML) models (e.g., loan approval or criminal sentencing). 
Without intervention, ML modes may discriminate against certain demographic groups (e.g., race, gender). 
For instance, Amazon developed a ML-based recruiting software that showed a strong bias against hiring women for technical jobs~\citep{amazon}.
\textit{Algorithmic fairness} research aims to develop algorithms that promote equitable treatment of 
different demographic groups
by correcting biases that may lead to unfair outcomes. 

A machine learning algorithm satisfies the \textit{demographic parity} fairness notion if the model predictions do not depend on the sensitive attributes~\citep{dwork2012fairness}. Demographic parity can compromise model performance, particularly when the true labels do depend on sensitive attributes. To address this limitation, \cite{hardt2016equality} introduced the concept of \textit{equalized odds}, which requires that the model predictions are conditionally independent of the sensitive attributes given the true labels. %

Two practical obstacles in the development of fair models are: (1) Training fair models requires access to \textit{sensitive data} (e.g., age, race, gender) in order to ensure fairness of predictions with respect to these attributes. However, data protection and privacy regulations (like E.U.'s General Data Protection Regulation and California's Consumer Privacy Act) restrict the usage of sensitive demographic consumer data. (2) Training data is often \textit{distributed} across different organizations, such as hospitals or banks, who may not share their data with third parties. 

To address obstacle (1), prior works~\citep{jagielski2019differentially,mozannar2020fair, tran2021differentially, tran2022sf, lgr23privatefair} have used 
\textit{differential privacy}~\citep{dwork2006calibrating}
to preserve the privacy of the sensitive data during fair model training. Informally, DP ensures that no adversary can infer much more about any individual piece of sensitive data than they could have inferred had that piece of data never been used. 
However, these approaches fail to address the second challenge, since they require access to the full centralized sensitive data. 

In this work, we address the two aforementioned challenges via fair private \textit{federated learning}~\citep{mcmahan2017originalFL} under an appropriate notion of differential privacy. 
Federated learning (FL) is a distributed learning framework in which silos collaborate to train a global model by exchanging focused updates, often with the orchestration of a central server. 
By permitting silos to collaborate without sharing their sensitive local data, FL offers an ideal solution to challenge (2). 

Although FL offers some privacy benefits to silos via local storage of data, this is not sufficient to prevent sensitive data from being leaked: model parameters or updates can leak data, e.g. via gradient or model inversion attacks~\citep{li2024analyzing,li2024exploring}. To prevent sensitive data from being leaked during FL, we will require the full transcript of silo $i$'s sent messages (e.g., local gradient updates) to be differentially private. This privacy requirement is known as \textit{inter-silo record-level differential privacy} (ISRL-DP)~\citep{lowy2023private,virginia}, defined formally in Section~\ref{sec: problem_setting}. For example, if the silos are hospitals, then ISRL-DP preserves the privacy of each patient's record, even if an adversary with server access colludes with the other hospitals to try to decode the data of hospital $i$.

\paragraph{Prior work.}
There is a lot of existing work on centralized private and fair learning~\citep{jagielski2019differentially, lgr23privatefair, tran2021differentially}, on private federated learning \citep{lowy2022private, lowy2021, pmlr-v130-girgis21a, gao2024privateheterogeneousfederatedlearning}, and on fair FL \citep{10.1609/aaai.v37i6.25911}. However, \textit{the literature on private and fair federated learning is very sparse}. In fact, the only related works we are aware of are due to~\citet{rodriguez2021enforcing, ling2024fedfdp}. The work of~\cite{rodriguez2021enforcing} does not prove any ISRL-DP guarantee for their algorithm, nor do they provide a convergence guarantee. The work of~\cite{ling2024fedfdp} only guarantees convergence for \textit{strongly convex} loss functions, limiting its applicability in ML; for example, linear/logistic regression and deep learning loss functions are not strongly convex. Moreover, the algorithm of~\cite{ling2024fedfdp} promotes a non-standard fairness notion known as ``balanced performance fairness,'' but not demographic parity or equalized odds. 

\paragraph{Contributions.} 
Motivated by the shortcomings of prior works, our work addresses the following question:
\begin{center}
\setlength{\fboxsep}{1pt} 
\fbox{
    \begin{minipage}{0.98\columnwidth}
   Can we develop an algorithm for fair and private federated learning that provably converges, even with loss functions that are not strongly convex?
    \end{minipage}
}

\end{center}

To answer this question, we develop a novel framework for promoting fairness and ISRL-DP with respect to sensitive attributes in a federated learning setting. Our approach builds on the centralized private fair learning method of \citet{lgr23privatefair}. Our framework is flexible, covering different fairness notions such as demographic parity and equalized odds.
Further, our algorithm provides:
\begin{enumerate}
    \item \textbf{Guaranteed ISRL-DP and convergence:} We prove that our ISRL-DP algorithm converges for any smooth (\textit{potentially non-convex}) loss function, even when mini-batches of data are used (i.e. stochastic optimization). Thus, our algorithm can be used in large-scale FL settings, where full batch training is not feasible. 
    \item \textbf{State-of-the-art empirical performance}: our ISRL-DP alogrithm achieves significantly improved fairness-accuracy tradeoffs on benchmark tasks across different privacy levels. For example, the equalized odds \textit{fairness violation of our algorithm was $95\%$ lower than the previous state-of-the-art}~\citep{ling2024fedfdp} for the same fixed accuracy level. Additionally, our algorithm even outperforms strong centralized DP fair baselines that do not provide the strong protection of ISRL-DP~\citet{tran2021differentially}. 
\end{enumerate}

As a byproduct of our algorithmic and analytical developments, we obtain the \textit{first convergence guarantee for distributed nonconvex-strongly concave min-max optimization under ISRL-DP} constraints. Moreover, our framework extends to \textit{hybrid centralization} settings, where some data features are centralized and others are decentralized (e.g., centralization of sensitive and decentralization of non-sensitive data).

\section{Problem Setting and Preliminaries}
\label{sec: problem_setting}

Let $Z = \{z_i = (x_i, s_i, y_i)\}_{i=1}^n  
$ be a data set with non-sensitive features $x_i \in \XX$, discrete sensitive attributes (e.g. race, gender) $s_i \in [k] \triangleq \{1, \ldots, k\}$, and labels $y_i \in [l] \triangleq \{1, \ldots, l\}$. Consider a federated learning setting with $N$ silos (e.g., hospitals or banks), each of which has data that is partitioned into sensitive and non-sensitive data divisions: $\{Z_j = (X_j, Y_j), S_j\}_{j=1}^N$, where $(X_j, Y_j) = \{x_{j,i}, y_{j,i}\}_{i=1}^{\tilde{n}}$, $S_j = \{s_{j,i}\}_{i=1}^{\tilde{n}}$, and $\tilde{n} = n/N \in \mathbb{N}$ is the number of local samples per silo.\footnote{Our algorithm and analysis readily extends to the case in which silo data sets contain different numbers of samples, via standard techniques (see e.g., \cite{lowy2023private})} Let $\haty_{\theta}(x)$ denote the model predictions  parameterized by $\theta$, and $\ell(\theta, x, y) = \ell(\haty_{\theta}(x), y)$ be a loss function (e.g. cross-entropy loss). Our goal is to (approximately) solve the empirical risk minimization (ERM) problem \begin{equation}
\label{eq: ERM}
    \min_{\theta}\left\{\widehat{\mathcal{L}}(\theta) := \frac{1}{N \tilde{n}} \sum_{j=1}^N \sum_{i=1}^{\tilde{n}} \ell(\theta, x_{ji}, y_{ji}) \right\}
\end{equation}
in a fair manner, while maintaining the differential privacy of the sensitive data $\{S_j\}_{j=1}^n$ under ISRL-DP. We consider two different notions of fairness in this work:\footnote{
Our method can also handle any other fairness notion that can be defined in terms of statistical (conditional) independence, such as equal opportunity. However, our method cannot handle all fairness notions: for example, false discovery rate and calibration error are not covered by our framework.
}

\begin{definition}[Fairness Notions]
Let $\Al: \ZZ \to \YY$ be a classifier.
\vspace{-.1cm}
\begin{itemize}
\vspace{-.3cm}
    \item $\Al$ satisfies \textit{demographic parity}~\citep{dwork2012fairness} if the predictions $\Al(Z)$ are statistically independent of the sensitive attributes.
    \vspace{-.15cm}
    \item $\Al$ satisfies \textit{equalized odds}~\citep{hardt2016equality} if the predictions $\Al(Z)$ are conditionally independent of the sensitive attributes given $Y = y$ for any $y \in [l]$. 
    \vspace{-.2cm}
\end{itemize}
\vspace{-.21cm}
\end{definition}
The choice of fairness notion depends on the application at hand (See \citet{10.1145/3376898} for discussion.) 

It has been demonstrated that achieving perfect fairness is \textit{impossible} for a differentially private algorithm that also achieves non-trivial accuracy~\citep{cummings}. Therefore, we focus on developing an algorithm that minimizes a certain measure of \textit{fairness violation} on the given dataset $Z$. Fairness violations can be quantified in various ways; see, \citet{dwork2012fairness, hardt2016equality,fermi} for an overview. As an example, if demographic parity is the desired fairness criterion, we can quantify the (empirical) demographic parity violation using the following measure: 
\begin{equation} 
    \label{eq: dem par violation} 
    \max_{\widehat{y} \in \mathcal{Y}} \max_{ s \in \mathcal{S}}\left|\hat{p}_{\hy|S}(\widehat{y}|s) - \hat{p}_{\hy}(\widehat{y})\right|,
\end{equation} 

where $\hat{p}$ represents the empirical probability computed from the data $(Z, \{\haty_i\}_{i=1}^n)$. Note that the demographic parity violation in~\eqref{eq: dem par violation} is zero if and only if demographic parity is satisfied.

Next, we define differential privacy.
Following the DP fair learning literature~\citep{jagielski2019differentially} and motivated by the discussion in the Introduction, we consider a relaxation of DP, in which only the \textit{sensitive attributes} require privacy.\footnote{However, the convergence guarantee of our algorithm easily extends to the case where privacy of the entire data set is needed.} In the centralized setting, we say $Z$ and $Z'$ are \textit{adjacent with respect to sensitive data} if $Z = \{(x_i, y_i, s_i)\}_{i=1}^n$, $Z' = \{(x_i, y_i, s'_i)\}_{i=1}^n$, and there is a unique $i \in [n]$ such that $s_i \neq s'_i$.  

\begin{definition}[Differential Privacy w.r.t. Sensitive 
Attributes]
\label{def: DP sens}
Let $\varepsilon \geq 0, ~\delta \in [0, 1).$ A randomized algorithm $\Al$ is \textit{$(\varepsilon, \delta)$-differentially private (DP) w.r.t. sensitive attributes S}  if for all pairs of data sets $Z, Z'$ that are \textit{adjacent w.r.t. sensitive attributes}, we have
\begin{equation}
\label{eq: DP}
\mathbb{P}(\Al(Z) \in O) \leq e^\varepsilon \mathbb{P}(\Al(Z) \in O) + \delta, 
\end{equation}
for all measurable sets $O \subseteq \YY$. 
\end{definition}

In the context of FL with $N$ silos, we say two distributed datasets $Z = (Z_1, \ldots, Z_N)$ and $Z' = (Z'_1, \ldots, Z'_N)$ with $Z_j = \{(x_{j,i}, y_{j,i}, s_{j,i})\}_{i=1}^{\tn}$ $Z'_j = \{(x'_{j,i}, y'_{j,i}, s'_{j,i})\}_{i=1}^{\tn}$ are adjacent if for every $j \in [N]$, there is at most one $i \in [\tn]$ such that $s_{j,i} \neq s'_{j, i}$. Thus, adjacent distributed datasets $Z$ and $Z'$ may differ in up to $N$ samples, one from each silo.

\begin{definition}[Inter-Silo Record-Level DP]
\label{def: ISRLDP}
A federated learning algorithm $\Al$ is $(\varepsilon, \delta)$-inter-silo-record-level DP (ISRL-DP) if, for each $j \in [N]$, the full transcript of silo $j$'s sent messages satisfies~\cref{eq: DP} for all adjacent distributed datasets $Z, Z'$ and any fixed settings of other silos’ data.
\vspace{-.03in}
\end{definition}
\vspace{-.03in}
By post-processing property of DP \citep{dworkrothbook}, Definition~\ref{def: ISRLDP} ensures that the  model parameters and the messages broadcast by the central server and are also DP.

As discussed in~\cref{sec: intro}, Definition \ref{def: DP sens} is useful if a company wants to train a fair model, but is unable to use the sensitive attributes collected in another silo (and is needed to train a fair model) due to privacy concerns and laws. Following~\cite{lgr23privatefair}, we shall impose the reasonably practical assumption that all data sets contain at least $\rho$-fraction of every sensitive attribute for some $\rho \in (0, 1)$.

\section{Private and Fair Federated ERM Framework}
\label{sec: privfairfederm}

A popular method in the literature to enforce fairness is to introduce a regularizer that penalizes the model for making unfair decisions~\citep{zhang2018mitigating, donini2018empirical, baharlouei2019rnyi}. Let $S$, $Y$, and $\Hat{Y}$ be the random variables corresponding to sensitive attributes, actual output, and predictions by the models. The regularization approach to fair ERM jointly optimizes for accuracy and fairness 
by solving
\[
\min_{\theta} \left\{\widehat{\mathcal{L}}(\theta) + \lambda \mathcal{D}(\hy, S, Y) \right\},
\]
where $\mathcal{D}$ is a measure of (conditional) statistical dependence (based on the fairness notion used) between the sensitive attributes $S$ and the predicted outputs $\Hat{Y}$. The dependency of $\mathcal{D}$ on $S$, $\hat{Y}$, and/or $Y$ varies for different fairness notions. For instance, for demographic parity, $\mathcal{D}$ just depends on $S$ and $\Hat{Y}$, while equalized odds $\mathcal{D}$ also depends on $Y$. The parameter $\lambda \geq 0$ controls the trade-off between accuracy and fairness. Inspired by the strong performance of \citet{fermi,lgr23privatefair}, we use variations of the $\chi^2$ divergence as our $\mathcal{D}$.

\begin{definition}[$\chi^2$ Divergence]
    The $\chi^2$ Divergence between two probability mass functions $P(x)$ and $Q(x)$ over the support of $X$ is defined as 
    \begin{align*}
        \chi^2(P||Q) = \sum_{x \in X} Q(x)\left(\frac{P(x)}{Q(x)} - 1\right)^2
    \end{align*}
\end{definition}

For demographic parity, we would ideally like to use $D_R(\hy, S) \triangleq \chi^2(p_{\hy,S}||p_{\hy} p_S)$ as our regularizer, where the \textit{true joint distribution} for the random variables $\hy$ and $S$ is given by $p_{\hy,S}$ and marginals are given by $p_{\hy}, p_S$, respectively. However, since the true distribution of $(\hy,S)$ is unknown in practice, we resort to an empirical estimate of the regularizer: $\widehat{D}_R(\hy, S) \triangleq \chi^2(\hat{p}_{\hy,S}||\hat{p}_{\hy}\hat{p}_S)$, where the empirical joint distribution for the random variables $\hy$ and $S$ is given by $\hat{p}_{\hy,S}$ and marginals by $\hat{p}_{\hy}, ~\hat{p}_S$ respectively. Similarly, for equalized odds, $D_R'(\hy, S) \triangleq \chi^2(p_{\hy,S|Y}||p_{\hy|Y} p_{S|Y})$, and we use $\widehat{D}'_R(\hy, S) \triangleq \chi^2(\hat{p}_{\hy,S|Y}||\hat{p}_{\hy|Y}\hat{p}_{S|Y})$ in practice. We write the full expressions of these regularizers in \cref{app:ermi expressions}. 

For concreteness, we consider demographic parity in what follows, but note that our developments extend easily to equalized oddds. Our approach to enforcing fairness is to augment~\cref{eq: ERM} with the $\chi^2$ regularizer and privately solve: 
\begin{equation}
\min_{\theta} 
\left
\{\text{FERMI}(\theta) := 
\widehat{\mathcal{L}}(\theta) +\lambda \widehat{D}_R(\hy_{\theta}(X), S) 
\right
\}.
\tag{FERMI obj.}
\label{eq: FERMI}
\end{equation}
The empirical divergence $\widehat{D}_R$ is an asymptotically unbiased estimator of population divergence $D_R$~\citep{fermi}, suggesting that solving ~\cref{eq: FERMI} should generalize well to the corresponding population risk minimization problem. 

The next question we address is: \textit{how do we solve~\eqref{eq: FERMI} in a distributed fashion, while satisfying ISRL-DP}? It is not obvious how to obtain statistically unbiased estimators of the gradients of $\widehat{D}_R(\hy_{\theta}(X), S)$ without directly computing $\nabla_{\theta} \widehat{D}_R(\hy_{\theta}(X), S)$ over the entire data set. But computing the gradient over the entire data set is not possible in the federated learning setting, since each silo stores its data locally in a decentralized manner. 

Fortunately, 
\cite{fermi} gives us a statistically unbiased estimator through a min-max problem formulation. For feature input $x$, let the predicted class labels be given by $\haty(x, \theta) = j \in [l]$ with probability $\FF_j(x, \theta)$, where $\FF(x, \theta)$ is differentiable in $\theta \in [0,1]^l$, and $\sum_{j=1}^l \FF_j(x, \theta) = 1$. For instance, $\FF(x, \theta) = (\FF_1(x, \theta), \ldots, \FF_l(x, \theta))$ could represent the output of a neural net after softmax layer or the probability label assigned by a logistic regression model. Then we have the following min-max re-formulation of~\cref{eq: FERMI}:

\begin{theorem}[\citet{fermi}]
\label{thm: informal Fermi as minmax}
There are differentiable functions $\widehat{\psi}_{ji}$ such that
\cref{eq: FERMI} is equivalent to
\small
\begin{align}
\label{eq: empirical minmax}
\min_{\theta} \max_{W \in \mathbb{R}^{k \times l}} 
\left\{
\widehat{F}(\theta, W) := 
\widehat{\mathcal{L}}(\theta) + \lambda \frac{1}{N \tilde{n}} \sum_{j=1}^N \sum_{i=1}^{\tilde{n}} \widehat{\psi}_{ji}(\theta, W)
\right\}.
\end{align}
\normalsize
Further, $\widehat{\psi}_{ji}(\theta, W)$ is strongly concave in $W$ for any $\theta$.
\end{theorem}

The functions $\widehat{\psi}_{ji}$ are given explicitly in~\cref{app: formal minmax}.
With \cref{thm: informal Fermi as minmax}, we can now claim that: for any batch on a particular silo $\mathcal{B}_j$ with size $m \in [\tilde{n}]$, the gradients (with respect to $\theta$ and $W$) of $\frac{1}{Nm} \sum_{j=1}^N\sum_{i \in \mathcal{B}_j} \ell(x_{ji}, y_{ji}; \theta) + \lambda \widehat{\psi}_{ji}(\theta, W)$ are statistically unbiased estimators of the gradients of $\widehat{F}(\theta, W)$, if $\mathcal{B}$ is drawn uniformly from $Z$. 
However, when differential privacy of the sensitive attributes is also desired, the formulation~\cref{eq: empirical minmax} presents some challenges, due to the non-convexity of $\widehat{F}(\cdot, W)$. \cite{lgr23privatefair} solve this problem in the centralized setting, but the proposed method may leak central data to the server and does not satisfy ISRL-DP.

Next, we develop our distributed ISRL-DP fair learning algorithm.

\subsection{ISRL-DP Fair Federated Learning via SteFFLe}
Our algorithm for privately solving the min-max FL problem~\eqref{eq: empirical minmax} is given in Algorithm~\ref{alg: steffle}. Algorithm~\ref{alg: steffle} is essentially a noisy distributed variation of \textit{stochastic gradient descent ascent} (SGDA). Gaussian noise is added to each silo's sensitive stochastic gradients $\nabla_{\theta} \widehat{\psi}, \nabla_{w} \widehat{\psi}$ to ensure ISRL-DP with respect to the sensitive attributes. Then, the server aggregates these noisy sensitive gradients and the noiseless non-sensitive gradients $\nabla_{\theta} \ell(x, y, \theta)$ and updates the model parameters $\theta_{t+1}$ and $W_{t+1}$ by taking descent and ascent steps. 

\begin{algorithm}[t]
    \caption{SteFFLe: Stochastic Private Fair Federated Learning}
    \label{alg: steffle}
	\begin{algorithmic}[1]
	 \STATE \textbf{Input}: $\{Z_j = \{x_{j,i}, y_{j,i}\}_{i=1}^{\tilde{n}}, \{s_{j,i}\}_{i=1}^{\tilde{n}}\}_{j=1}^N$, $\theta_0 \in \mathbb{R}^{d_{\theta}}, ~W_0 = 0 \in 
	 \mathbb{R}^{k \times l}$, step-sizes $(\eta_\theta, \eta_w),$ 
	 fairness parameter $\lambda \geq 0,$ iteration number $T$, minibatch size $|B_t| = m \in [\tilde{n}]$, set $\WW \subset 
	 \mathbb{R}^{k \times l}$, noise parameters $\{\sigma_{j,w}^2, \sigma_{j,\theta}^2\}_{j=1}^N$. 
	 \STATE  Compute $\widehat{P}_{S}^{-1/2} = {\rm diag}({\widehat{p}}_{S}(1)^{-1/2}, \ldots, \widehat{p}_{S}(k)^{-1/2})$, where $\widehat{p}_{S}(r) := \frac{1}{N\tilde{n}}\sum_{j=1}^N\sum_{i=1}^{\tilde{n}} \mathbbm{1}_{\{s_{ji} = r \}} \geq \rho > 0$. 
	    \FOR {$t = 0, 1, \ldots, T-1$}
	    \STATE Central server sends $\theta_t, W_t$ to all silos.
	    \FOR {$j \in [N]$ \textbf{in parallel}} 
	    \STATE Silo $j$ draws a mini-batch $B_t$ of data points $\{(x_{j,i}, y_{j,i}), s_{j,i}\}_{i\in B_t}$. 
	    \STATE Silo $j$'s non-sensitive division computes stochastic gradient $g_{t, j} := \frac{1}{|B_t|} \sum_{i \in B_t} \nabla_{\theta} \ell(x_{j,i}, y_{j,i}, \theta_t)$ and sends $\{ \FF(x_{j,i}, \theta_t), \nabla \FF(x_{j,i}, \theta_t), j, i\}_{i \in B_t}$ to sensitive data division. 
	    \STATE Silo $j$'s sensitive division computes noisy sensitive stochastic gradients $h_{t, j, \theta} := \frac{1}{|B_t|} \sum_{i \in B_t} \nabla_{\theta} \widehat{\psi}_{j,i}(\theta_t, W_t) + u_{t,j}$ and $h_{t, j, w} := \frac{1}{|B_t|} \sum_{i \in B_t} \nabla_{w} \widehat{\psi}_{j,i}(\theta_t, W_t) + V_{t,j}$, where $u_{t,j} \sim \mathcal{N}(0, \sigma_{j,\theta}^2 \mathbf{I}_{d_\theta})$ and $V_{t,j}$ is a $k \times l$ matrix with independent random Gaussian entries $(V_t)_{q, r} \sim \mathcal{N}(0, \sigma_{j,w}^2)$.
        \STATE Silo $j$ broadcasts $g_{t,j}$, $h_{t,j,\theta}$, and $h_{t,j,w}$ to the central server.
	    \ENDFOR
	    \STATE Central server updates \small$\theta_{t+1} \gets \theta_t -  \frac{\eta_\theta}{N} \sum_{j=1}^N [g_{t,j} + \lambda h_{t,j, \theta}]$\normalsize ~and \small$
	        W_{t+1} \gets 
	        \Pi_{\WW} 
	        \Big(
	        W_t + 
	        \frac{\lambda \eta_w}{N} \sum_{j=1}^N h_{t,j, w} 
	   \Big)$\normalsize.
		\ENDFOR
		\STATE Pick $\hat{t}$ uniformly at random from $\{1, \ldots, T\}.$\\
		\STATE \textbf{Return:} $\hat{\theta}_T := \theta_{\hat{t}}.$
	\end{algorithmic}
\end{algorithm}

\begin{theorem}
\label{thm: steffle privacy}
    Let $\varepsilon \leq 2\ln(1/\delta)$, $\delta \in (0,1)$, and $T \geq \left(\tn \frac{\sqrt{\varepsilon}}{2 |B_t|}\right)^2$. Assume $\FF(x, \cdot)$ is $L_\theta$-Lipschitz for all $x$, and $|(W_t)_{r,q}| \leq D$ for all $t \in [T], r \in [k], q \in [l]$.  Then, for %
    $\sigma_{j, w}^2 \geq \frac{16 T \ln(1/\delta)}{\varepsilon^2 \tn^2 \rho}$ and $\sigma_{j,\theta}^2 \geq \frac{16 L_\theta^2 D^2 \ln(1/\delta) T}{\varepsilon^2 \tn^2 \rho}$, \cref{alg: steffle} is $(\varepsilon, \delta)$-ISRL-DP with respect to the sensitive attributes for all data sets containing at least $\rho$-fraction of minority attributes.
\end{theorem}

See~\cref{app: proof of privacy steffle} for the proof. Next, we provide a convergence guarantee for \cref{alg: steffle}.

\begin{theorem}
\label{thm: steffle utility}
    Assume that the loss function $\ell(\cdot, x, y)$ is Lipschitz and $\ell(\cdot, x, y)$ and $\FF(x, \cdot)$ have Lipschitz gradients. Then, there exist algorithmic parameters such that~\cref{alg: steffle} returns $\hat{\theta}_T$ with \[
\expec\|\nabla \text{FERMI}(\hat{\theta}_T)\|^2 = \mathcal{O}\left(\frac{\sqrt{\max(d_{\theta}, kl) \ln(1/\delta)}}{\varepsilon \tn \sqrt{N}}\right).
\]
\end{theorem}

Compared to the central DP stationarity gap bound obtained in~\citet{lgr23privatefair} (with $n = \tilde{n} N$), the bound in~\ref{thm: steffle utility} is larger by a factor of $\sqrt{N}$.
This is because ISRL-DP is a stronger privacy notion than central DP~\citep{lowy2023private}.

The proof of~\cref{thm: noisy fed SGDA convergence} follows from careful tracking of noise variance and sampling of data obtained from the different silos. A key observation is that even though the sampling is distributed across silos, the expected value of gradient after this modified form of sampling is an unbiased estimator of the global loss function due to linearity of expectation. Moreover, by averaging silos' noisy gradients, we reduce the total privacy noise variance.  
The rest of the proof leverages the DP min-max optimization techniques of~\citet{lgr23privatefair}. See~\cref{app: proof of fed-fermi} for the detailed proof. In fact, in~\cref{app: proof of fed-fermi}, we prove~\cref{thm: noisy fed SGDA convergence}, which is a general result that applies to \textit{all smooth non-convex strongly-concave min-max optimization problems}, being of independent interest to the private optimization and federated learning community.

In Algorithm~\ref{alg: steffle}, we implicitly assume that the \textit{frequency} of each sensitive attribute is known in order to compute $\Hat{P_S}$ and broadcast it to the silos. This assumption is not very restrictive: In practice, releasing the frequency of the sensitive attributes of data is very common. Moreover, it is straightforward to privately estimate $\Hat{P_S}$ using DP histograms. 
Thus, for simplicity, we assumed $\Hat{P_S}$ to be known.

In the next section, we show our framework extends to \textit{hybrid} centralization settings. 

\section{Different Modes of Data Centralization}
\label{sec: realworld}
\newcommand{\andy}[1]{\textcolor{red}{[Andy: #1]}}
Recall that we have assumed each silo is divided into two distinct parts: one that holds the \textit{sensitive} data and another that holds \textit{non-sensitive} data. The two divisions within each silo can communicate with each other and with all the sensitive and non-sensitive divisions of other silos. 
Leveraging this subtlety, we show how to model a wide range of hybrid centralized/distributed learning tasks that involve privacy of sensitive attributes. We will illustrate how Algorithm~\ref{alg: steffle} readily extends to these hybrid tasks.

\textbf{One Silo, Centralized Sensitive and Non-Sensitive Data in Separate Subdivisions.}
An example of this can be seen in healthcare organizations where the sensitive part of data can only be accessed by authorized personnel.
In this case, we have $N = 1$ silo in SteFFLe. The updates from the sensitive subdivision are private due to the ISRL-DP guarantee in~\cref{thm: steffle privacy}. Theorem~\ref{thm: steffle utility}  recovers the stationarity bound in~\citet{lgr23privatefair}.  

\textbf{Centralized Sensitive Data and Decentralized Non-Sensitive Data.}
For example, the United States Census Bureau is a silo containing centralized sensitive data; the non-sensitive features are distributed across different silos (e.g., banks or healthcare providers that use Census data).
In this case, we have $1$ silo containing sensitive features and $N$ silos containing non-sensitive features.
Our algorithm can be used to train models in this setting: in round $t$, instead of querying silo $i$'s sensitive division, the central server queries the central sensitive silo and receives noisy ISRL-DP sensitive gradients. These noisy sensitive gradients are combined with the noiseless non-sensitive gradients from each of the non-sensitive silos, and then the model is updated.  
See~\cref{examples:hybrid} for more details.

\begin{figure*}[t!]
        \subfloat[$\varepsilon = 1$, \textit{Homogeneous} ($h = 0$)]{
            \includegraphics[width=.48\linewidth]{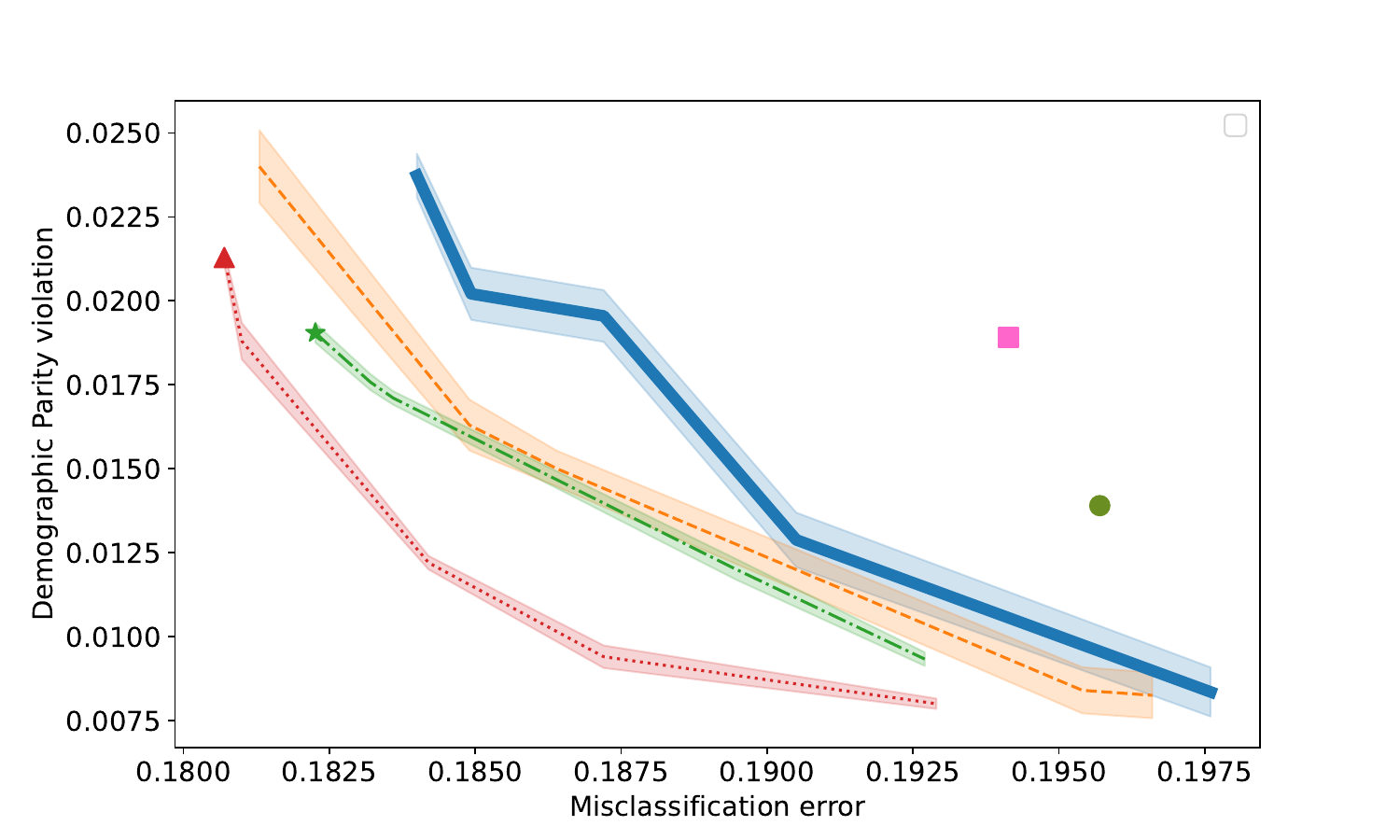}
            \label{subfig:a1}
        }\hfill
        \subfloat[$\varepsilon = 1$, \textit{Heterogeneous} ($h = 0.75$)]{
            \includegraphics[width=.48\linewidth]{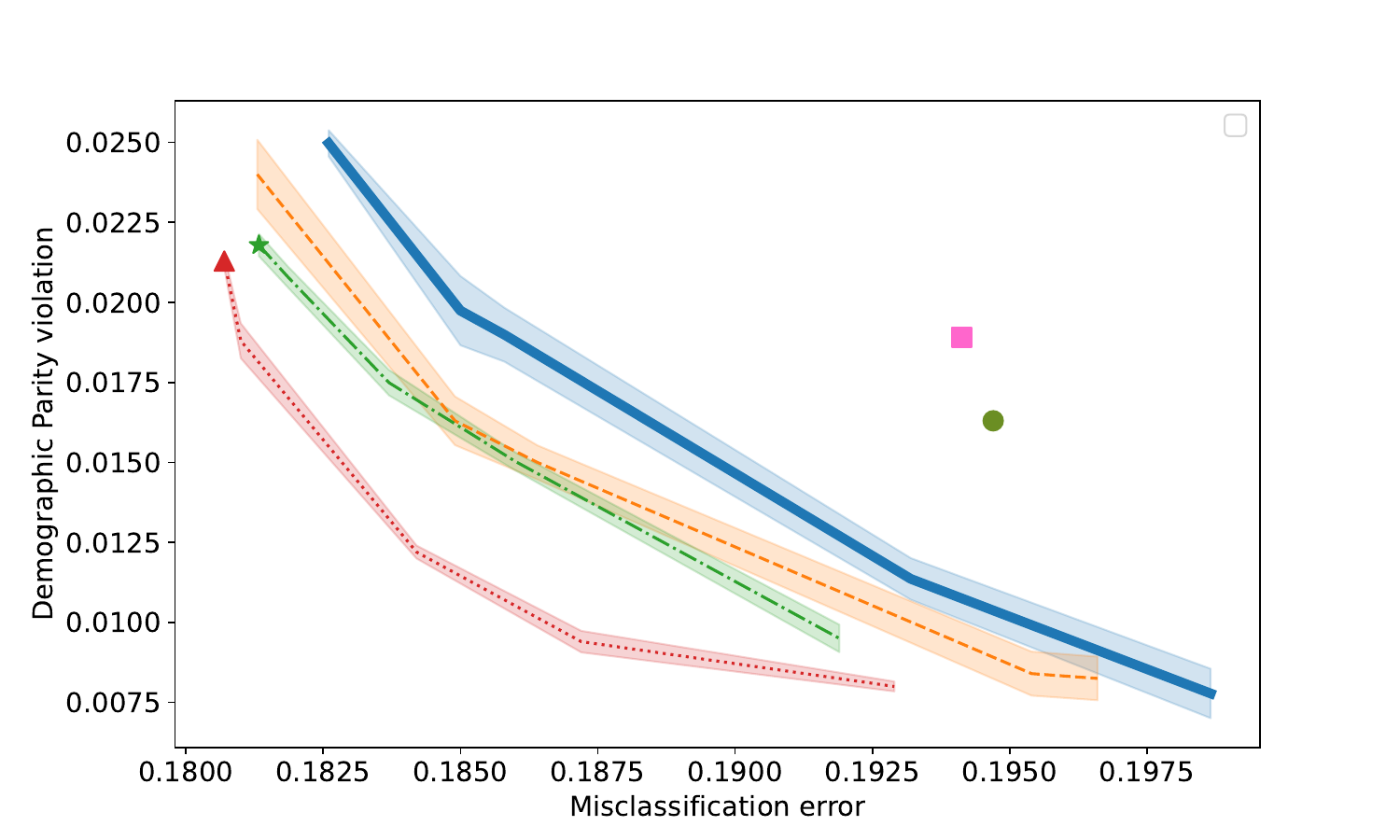}
            \label{subfig:b1}
        }\\
        \subfloat[$\varepsilon = 3$, \textit{Homogeneous} ($h = 0$)]{
            \includegraphics[width=.48\linewidth]{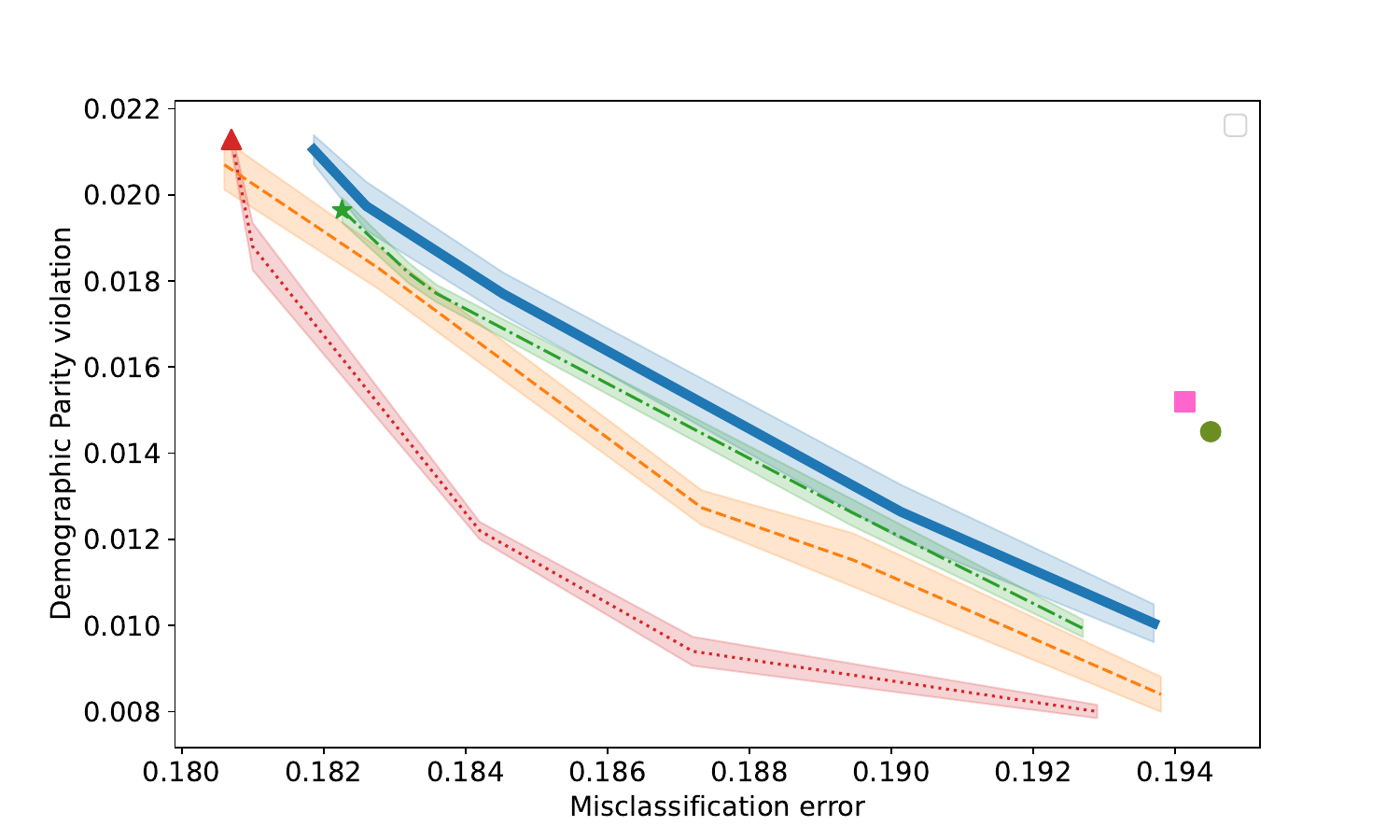}
            \label{subfig:c1}
        }\hfill
        \subfloat[$\varepsilon = 3$, \textit{Heterogeneous} ($h = 0.75$)]{
            \includegraphics[width=.48\linewidth]{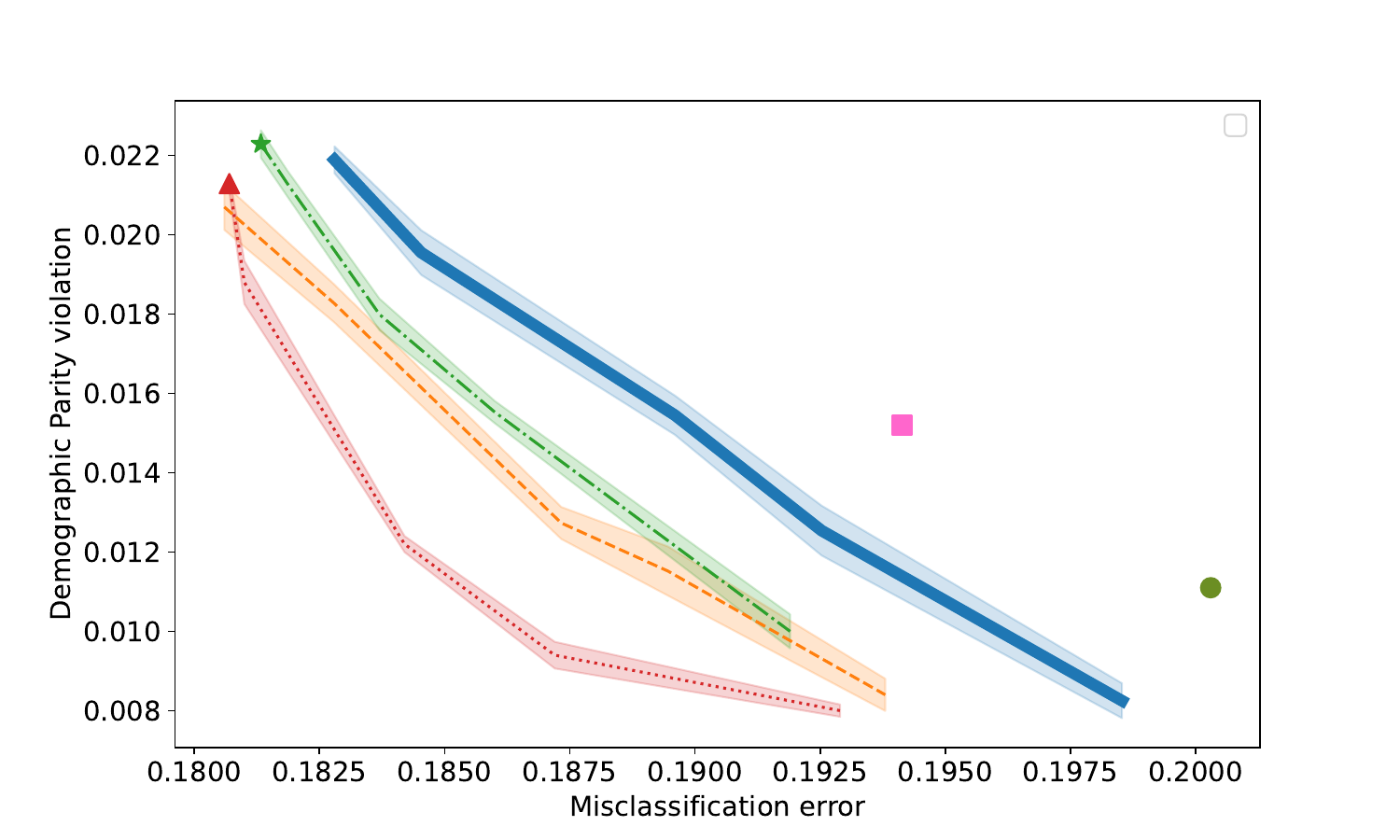}%
            \label{subfig:d1}
        }\\
        \begin{center}
            \subfloat[Plot Legend]{
            \includegraphics[width=0.50\linewidth]{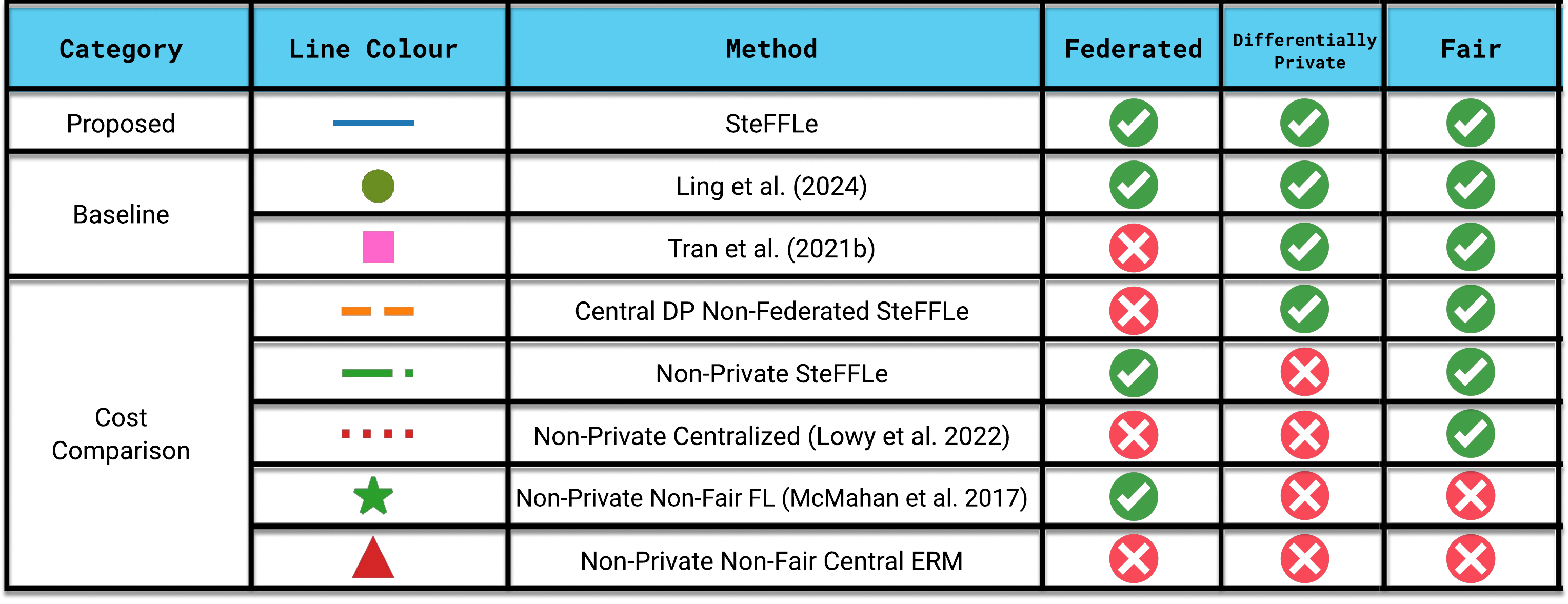}%
            \label{subfig:e1}
        }
        \end{center}

        \caption{Demographic parity vs Misclassification error on \textit{Credit Card} dataset (\textit{Number of Silos} = 3)}
        \label{fig:fig_cc_main}
\end{figure*}

\paragraph{General Case: Arbitrary Numbers of Sensitive and Non-Sensitive Silos.}
In~\cref{examples:hybrid}, we explain how to extend our algorithm to the completely general hybrid centralization setting with an arbitrary numbers of sensitive and non-sensitive silos.

\begin{figure*}[t!]
        \subfloat[$\varepsilon = 1$, \textit{Number of Silos} = 3, Varying $h$]{
            \includegraphics[width=.48\linewidth]{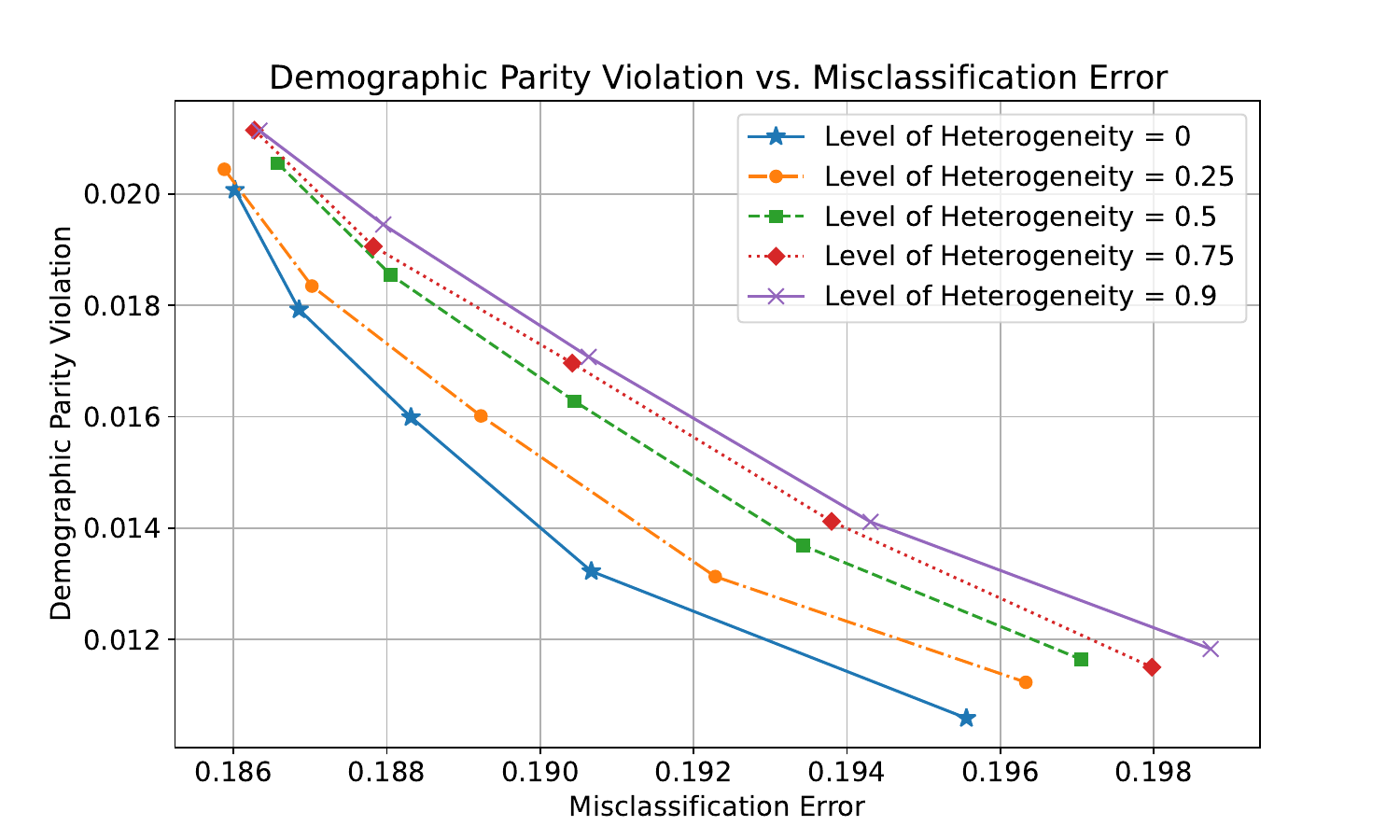}
            \label{subfig:a2}
        }\hfill
        \subfloat[$\varepsilon = 3$, \textit{Number of Silos} = 3, Varying $h$]{
            \includegraphics[width=.48\linewidth]{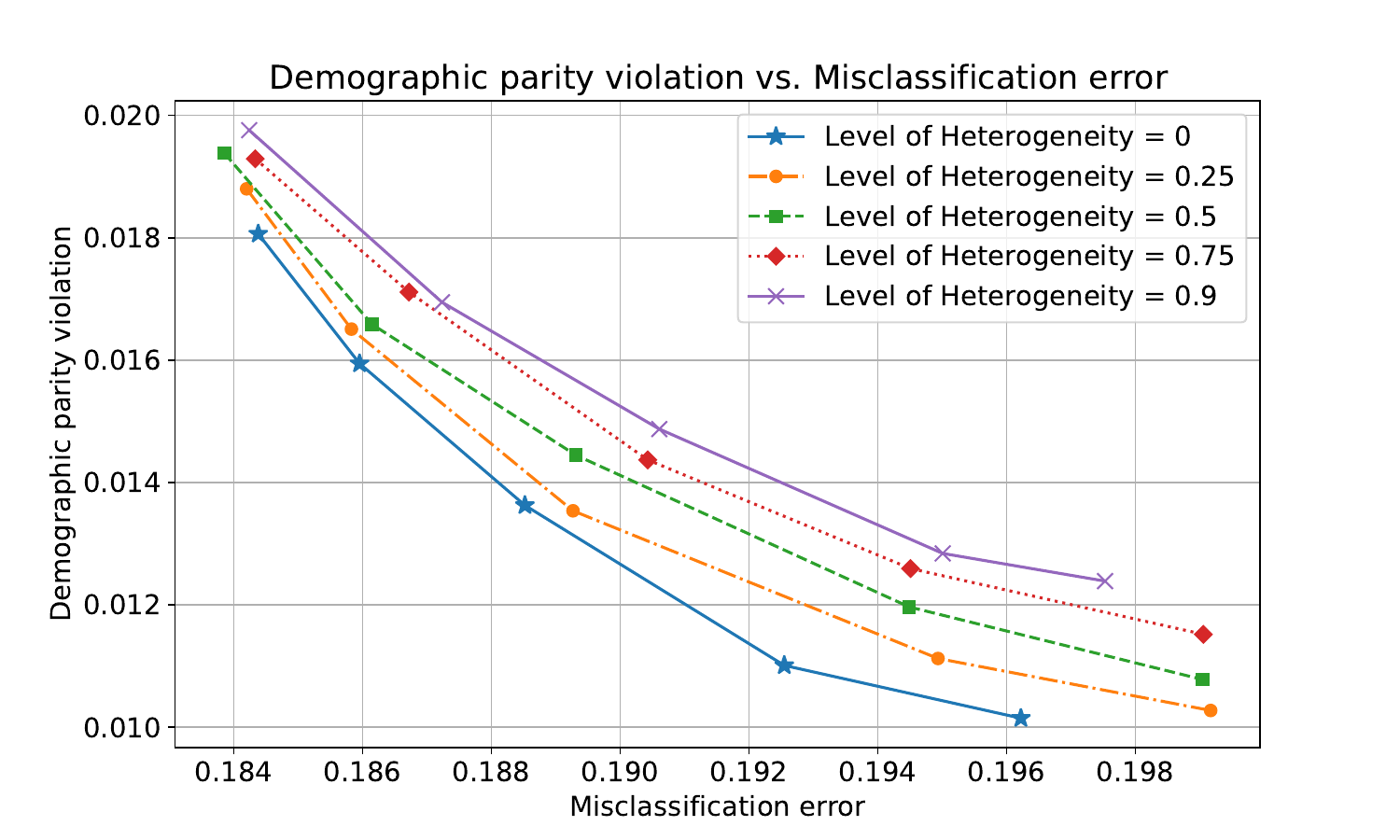}
            \label{subfig:b2}
        }\\
        \subfloat[$\varepsilon = 1$, \textit{Homogeneous} ($h = 0$), Varying $N$]{
            \includegraphics[width=.48\linewidth]{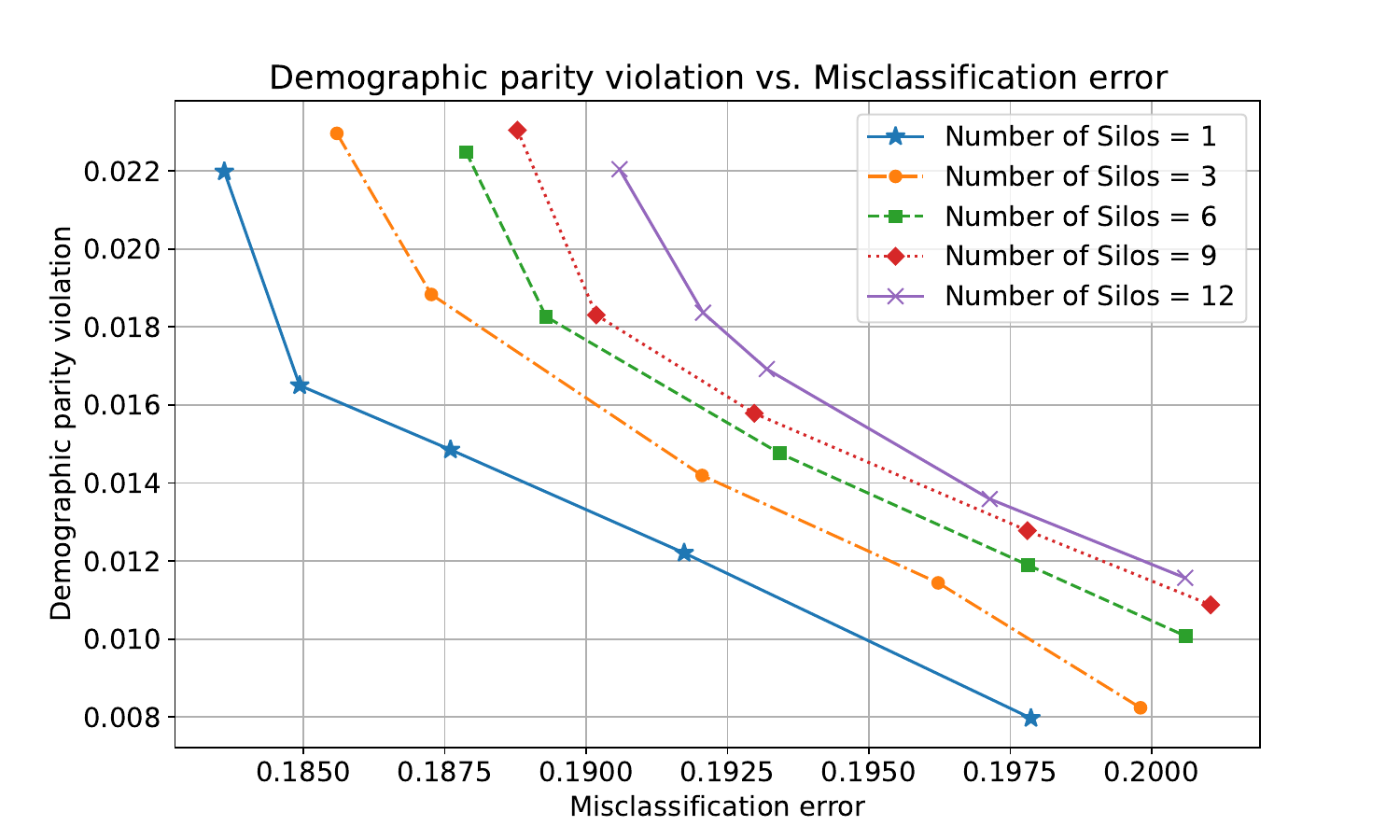}
            \label{subfig:c2}
        }\hfill
        \subfloat[$\varepsilon = 1$, \textit{Heterogeneous} ($h = 0.75$), Varying $N$]{
            \includegraphics[width=.48\linewidth]{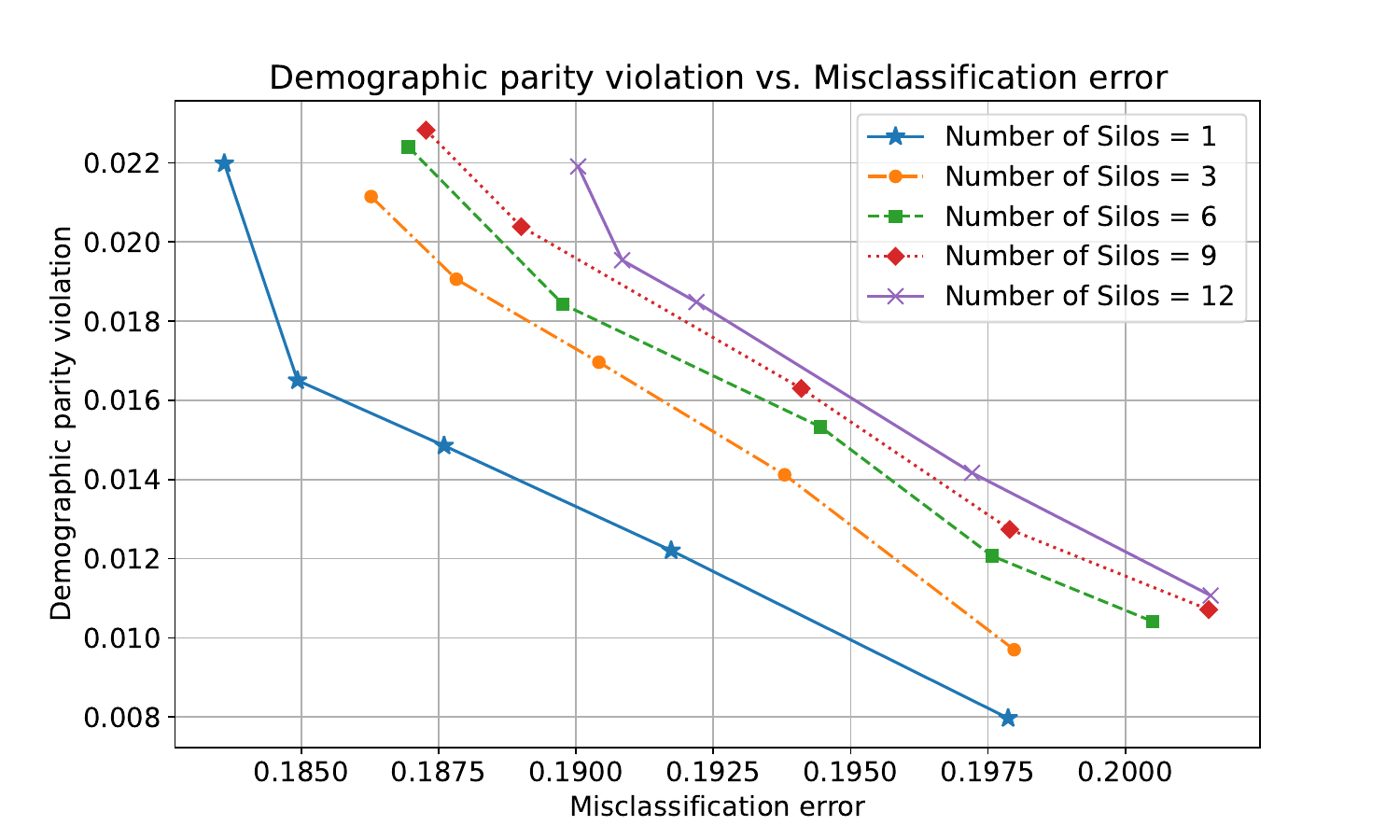}%
            \label{subfig:d2}
        }
        \caption{Varying Levels of Heterogeneity ($h$) and Number of Silos ($N$) on the \textit{Credit Card} dataset}
        \label{fig:fighetero_silo_var}
\end{figure*}

\section{Numerical Experiments}
\label{sec: experiments}
In this section, we evaluate the performance of our algorithm in terms of fairness violation vs. test error for different levels of privacy, levels of silo heterogeneity, and numbers of silos. We present our results in two parts:
In Section ~\ref{ssec:tabdata}, we assess the performance of our method in training logistic regression models on several benchmark tabular datasets. %
In Section \ref{sssec:heterosilo}, we discuss how the fairness-accuracy tradeoffs are affected by silo heterogeneity and by the number of silos for a fixed privacy level.

\paragraph{Average results.} To evaluate the overall performance of our algorithm and the existing baselines, we calculated the performance gain with respect to fairness violation (for fixed accuracy level) that our model yields \textit{over all the datasets}. \ul{We obtained reductions in demographic parity violations of around 75.47\% and 52.93\%} compared with \citet{tran2021differentially} and \cite{ling2024fedfdp}. Note that the algorithm of~\citet{tran2021differentially} is \textit{not ISRL-DP}, instead satisfying only the weaker notion of central DP. 
We also obtained an average \ul{reduction in equalized odds violation of 95.42\%} compared to \cite{ling2024fedfdp}.
We specify our experimental setup, datasets, methods and additional results that we compare against in \cref{app: experiments}.

\subsection{Federated, Private, and Fair Logistic Regression}
\label{ssec:tabdata}
In the first set of experiments we train a logistic regression model using SteFFLe (Algorithm~\ref{alg: steffle}) to promote demographic parity. We compare SteFFLe against all applicable publicly available baselines in each experiment. We carefully tuned the hyperparameters of all baselines for fair comparison. We find that \textit{SteFFLe consistently outperforms all state-of-the-art baselines across all data sets in all privacy and heterogeneity levels}.  

\paragraph{Baselines.} The baselines include: (1) the approach by \citet{tran2021differentially}, which is \textit{central} differentially private and fair but \textit{not federated and not ISRL-DP}; (2) the method of \citet{ling2024fedfdp}, which incorporates federated learning, ISRL-DP, and fairness. These were the only DP fair baselines with code made publicly available for each experiment. 

Additionally, we examine the \textit{cost of incorporating federated learning and ISRL-DP} by measuring the fairness-accuracy trade-offs for different \textit{variations of SteFFLE}. These variations include: \textit{Central DP SteFFLe}~\cite{lgr23privatefair}, which is not ISRL-DP or federated, but still satisfies the weaker central DP notion and still promotes fairness; \textit{Non-Private SteFFLe}, which is fair and federated, but not private; \textit{Non-Private Centralized}~\cite{fermi}, which is fair, but not private or federated; \textit{Non-Private Non-Fair FL}~\cite{mcmahan17}, which uses federated averaging; and \textit{Non-Private Non-Fair Central ERM}, which simply uses SGD. See Figure~\ref{fig:fig_cc_main} and the legend therein for our results on \textit{Credit Card} dataset.

\paragraph{Datasets.} We use three benchmark tabular datasets: Credit-Card, Adult Income, and Retired Adult  dataset from the UCI machine learning repository~(\citet{Dua:2019}). The predicted variables and sensitive attributes are both binary in these datasets. We analyze fairness-accuracy trade-offs with three different privacy budgets $\varepsilon \in \{1, 3, 9\}$ and two different values of heterogeneity levels $h  = 0$ (homogeneous setting) and $h = 0.75$ (heterogeneous setting), keeping the number of silos $N = 3$ for each dataset. We compare against state-of-the-art algorithms proposed in~\citet{ling2024fedfdp} and (the demographic parity objective of) \citet{tran2021differentially}. The results displayed are averages over 15 trials (random seeds) for each value of $\varepsilon, h$ and $N$.

\paragraph{Results for different datasets.}
Selected results for private and fair federated logistic regression on the Credit Card dataset are shown in \cref{fig:fig_cc_main}. The remaining results of the Credit Card dataset and experiments of Adult and Retired Adult dataset are shown in \cref{ssec:demopair_app} and \cref{ssec:eo_app}. For logistic regression with equalized odds as the fairness violation, we provide further results (for a modified version of SteFFLe) on the Credit Card dataset in~\cref{ssec:eqodds_steffle_variant}. Compared to the baselines~\cite{tran2021differentially} and~\cite{ling2024fedfdp}, SteFFLe offers superior fairness-accuracy tradeoffs at all privacy ($\varepsilon$) and heterogeneity levels ($h$) across all three datasets. Moreover, the method of~\cite{tran2021differentially} is not ISRL-DP.  

\subsection{Impact of Silo Heterogeneity and the Number of Silos}
\label{sssec:heterosilo}

In this section, we analyze the impact on SteFFLe's performance due to \textit{varying heterogeneity levels} and \textit{the number of silos} on the fairness-error trade-off, with a fixed privacy budget of $\varepsilon = 1$. We analyze how these factors affect demographic parity violation and misclassification error on the Credit Card dataset, as depicted in ~\cref{fig:fighetero_silo_var}. 

\paragraph{Heterogeneous silo data is challenging in private fair FL.} We conducted experiments with silo heterogeneity levels ranging from $0$ to $0.9$, with $0$ being homogeneous and $1$ being heterogeneous. In \hyperref[fig:fighetero_silo_var]{Fig.2(a) and 2(b)}, the results demonstrate a \textit{clear increase in both misclassification error and demographic parity violation as heterogeneity increases}, for a fixed number of $N=3$ silos. This indicates that higher silo heterogeneity exacerbates the model’s difficulty in achieving an optimal balance between fairness and accuracy.

\hyperref[fig:fighetero_silo_var]{Fig. 2(c) and 2(d)} illustrates the effect of the number of silos 
on performance in both the homogeneous and heterogeneous settings. We vary the number of silos between $N \in [1, 12]$. In the homogeneous settings, as the number of silos increases from 1 to 12, both demographic parity violation rise and misclassification error grows. A similar trend is apparent in the heterogeneous setting, where an increase in the number of silos results in a proportional rise in both demographic parity violation and misclassification error in \hyperref[fig:fighetero_silo_var]{Fig. 2 (d)}. These findings suggest that \textit{increasing the number of silos amplifies the challenges of maintaining fairness and accuracy, particularly under federated learning frameworks which incorporate privacy constraints}.

\section{Conclusion and Discussion}
Motivated by pressing ethical and legal concerns, we considered the problem of training fair and private ML models with decentralized data. We developed an algorithm that satisfies the strong ISRL-DP guarantee. 
We proved that our ISRL-DP algorithm converges for any minibatch size, without requiring (strong) convexity of the loss function. Finally, numerical experiments on several benchmark fairness data sets demonstrated that our method offers substantial fairness-accuracy benefits over the prior art, across different levels of privacy and silo heterogeneity. Our experiments also highlighted the challenges of silo heterogeneity for fair and accurate ISRL-DP FL. 

Given the practical importance of private and fair federated learning, we hope that our paper inspires future researchers to continue working in this area. On the theoretical front, it would be interesting to understand \textit{fundamental} tradeoffs between ISRL-DP and fairness and accuracy in a similar vein to the works of~\cite{cummings, agarwal2021trade} for the centralized setting. Practically, there may be room for further improvement over our algorithm, given the test error gap between our method and the non-private and non-fair baselines, particularly in the heterogeneous setting.

\bigskip
\noindent\textbf{Acknowledgments} \\ \\
\noindent
This work was supported in part with funding from the NSF CAREER award 2144985, from the YIP AFOSR award, a gift from the USC-Meta Center for Research and Education in AI \& Learning, and a gift from Google. Andrew Lowy’s research is supported by NSF grant 2023239 and the AFOSR award FA9550-21-1-0084.

\bibliographystyle{unsrtnat}
\bibliography{template}  

\appendix
\onecolumn

\section{Demographic Parity and Equalized Odds Version of ERMI}
\label{app:ermi expressions}
If demographic parity~\citep{dwork2012fairness} is the desired fairness notion, then one should use the following definition of Chi-Squared divergence as a regularizer~\citet{fermi}: 
\begin{align}
\widehat{D}_R(\hy, S) &:=
\sum_{j \in [l]} \sum_{r \in [k]} \frac{\hat{p}_{\hy, S}(j, r)^2}{\hat{p}_{\hy}(j) \hat{p}_S(r)} - 1
\label{eq: ERMI}
\end{align}
 
For equalized odds~\citep{hardt16}, one should use the following expression as a regularizer~\citet{fermi}: 
\begin{align}
    \widehat{D}_R(\hy; S |Y) &:= \mathbb{E}\left\{\frac{\hat{p}_{\hy, S|Y}(\hy, S|Y) }{\hat{p}_{\hy|Y}(\hy|Y) \hat{p}_{S|Y}(S|Y)}  \right\} - 1\nonumber \\
 &= \sum_{y = 1}^l \sum_{j=1}^l \sum_{r=1}^k \frac{\hat{p}_{\hy, S |Y} (j, r | y)^2}{\hat{p}_{\hy|Y}(j|y) \hat{p}_{S|Y}(r|y)}\hat{p}_{Y}(y)  - 1.
 \end{align}
 In particular, if $D_R(\hy; S |Y) = 0$, then $\hy$ and $S$ are conditionally independent given $Y$ (i.e. equalized odds is satisfied).

\section{SteFFLe Algorithm: Privacy}
\label{app: proof of privacy steffle}
To prove~\cref{thm: steffle privacy}, we first consider the following definitions. For a particular silo $j$ at a particular iteration t, let the batch be denoted by $|B_t|$ and $\Delta^j_w$ denote the L2 sensitivity of gradient updates in silo $j$ with respect to $\theta$ and $w$ respectively. 
\[\Delta^j_\theta = 
\sup_{Z_j \sim Z'_j, \theta, W} \left\|\frac{1}{m}\sum_{i \in B_t}  \left[\nabla_{\theta} \widehat{\psi}(\theta, W; z_{ji}) -  \nabla_{\theta} \widehat{\psi}(\theta, W; z'_{ji})\right]\right\|,
\]
where we write $Z_j \sim Z_j'$ to mean that $Z_j$ and $Z_j'$ are two data sets (both with $\rho$-fraction of minority attributes) that differ in exactly one person's sensitive data. Likewise,
\[\Delta^j_W = 
\sup_{Z_j \sim Z'_j, \theta, W} \left\|\frac{1}{m}\sum_{i \in B_t}  \left[\nabla_{W} \widehat{\psi}(\theta, W; z_{ji}) -  \nabla_{W} \widehat{\psi}(\theta, W; z'_{ji})\right]\right\|.
\]

To upper bound the sensitivity of the above quantities, we use the following result from~\cite{lgr23privatefair}.

\begin{lemma}[\cite{lgr23privatefair}]
\label{lem:sensitivity}
    With the above definition of neighbouring databases $Z$ and $Z'$, for $B \subset [n]$, 
    \[
    \sup_{Z \sim Z', \theta, W} \left\|\frac{1}{|B|}\sum_{i \in B}  \left[\nabla_{\theta} \widehat{\psi}(\theta, W; z_{i}) -  \nabla_{\theta} \widehat{\psi}(\theta, W; z'_{i})\right]\right\| \leq \frac{8D^2L_\theta^2}{|B|^2 \rho},
    \]
    and 
    \[
    \sup_{Z \sim Z', \theta, W} \left\|\frac{1}{|B|}\sum_{i \in B}  \left[\nabla_{W} \widehat{\psi}(\theta, W; z_{i}) -  \nabla_{W} \widehat{\psi}(\theta, W; z'_{i})\right]\right\| \leq \frac{8}{|B|^2 \rho},
    \]
\end{lemma}

\begin{proof}[Proof of~\cref{thm: steffle privacy}]
    Since, we have to guarantee ISRL-DP, we require that the gradients broadcasted from each \textit{silo} is private. Using the fact that Lemma~\ref{lem:sensitivity} holds for any silo, we get that $\Delta^j_\theta \leq \frac{8D^2L_\theta^2}{|B_t|^2 \rho}$ and $\Delta^j_W \leq \frac{8}{|B_t|^2 \rho}$. By the Moments Accountant Theorem 1 of \cite{abadi16} and the fact that each silo has a total of $\tilde{n}$ datapoints, the claim holds.
\end{proof}

\section{SteFFLe Algorithm: Utility}
\label{app: proof of fed-fermi}

To prove~\cref{thm: steffle utility}, we will first derive a more general result. Namely, in~\cref{subsec: Fed SGDA}, we will provide a precise upper bound on the stationarity gap of noisy ISRL-DP federated stochastic gradient descent ascent (ISRL-DP-Fed-SGDA). We build on \cite{lgr23privatefair} to derive the stationarity gap.

\subsection{Noisy ISRL-DP Fed-SGDA for Nonconvex-Strongly Concave Min-Max FL Problems}
\label{subsec: Fed SGDA}
In this subsection, we will prove~\cref{thm: noisy fed SGDA convergence}, which implies the utility guarantee claimed in~\cref{thm: steffle utility} (which we re-state as a standalone theorem in~\cref{thm: fed FERMI utility}).

Let $Z = (Z_1, \ldots, Z_N)$ be a distributed dataset with $Z_j = \{z_{j,i}\}_{i=1}^{\tilde{n}}$ for $j \in [N]$. 
Consider a generic (smooth) nonconvex-strongly concave min-max federated ERM problem: 
\small
\begin{equation}
\label{eq: minmax2}
    \min_{\theta \in \rdt} \max_{w \in \WW} \left\{F(\theta, w) := \frac{1}{N \tilde{n}} \sum_{j=1}^N \sum_{i=1}^{\tilde{n}} f(\theta, w; z_{j,i})\right\},
\end{equation}
\normalsize
where $f(\theta, \cdot; z)$ is $\mu$-strongly concave for all $\theta, z$ but
$f(\cdot, w; z)$ is potentially non-convex. Grant~Assumption~\ref{ass: smooth}.  
\begin{algorithm}
    \caption{Noisy ISRL-DP Federated Stochastic Gradient Descent-Ascent (ISRL-DP-Fed-SGDA)
    }
    \label{alg: noisy Fed SGDA}
	\begin{algorithmic}[1]
	 \STATE \textbf{Input}: data $Z$, $\theta_0 \in \mathbb{R}^{d_{\theta}}, ~w_0 \in \WW$, step-sizes $(\eta_\theta, \eta_w),$ privacy noise parameters $\tilde{\sigma_\tth}, \tilde{\sigma_w}$, batch size $m$, iteration number $T \geq 1$. 
	    \FOR {$t = 0, 1, \ldots, T-1$}
	    \STATE Central server communicates $(\theta_t, w_t)$ to silos. 
	    \FOR {$j \in [N]$ \textbf{in parallel}}
	    \STATE Silo $j$ draws a fresh batch of data points $\{z_{j,i}\}_{i=1}^m$ uniformly at random from $Z_j$ with replacement. 
	    \STATE Silo $j$ draws fresh independent Gaussian noises $u_j \sim \mathcal{N}(0, \tilde{\sigma}_{\theta}^2 \mathbf{I}_{d_{\theta}})$ and $v_j \sim  \mathcal{N}(0, \tilde{\sigma}_{w}^2 \mathbf{I}_{d_{w}})$.
	    \STATE Silo $j$ communicates noisy stochastic gradients $H_{j, \theta} = \frac{1}{m} \sum_{i=1}^m \nabla_{\theta} f(\theta_t, w_t, z_{j,i}) + u_j$ and $H_{j, w} = \frac{1}{m} \sum_{i=1}^m \nabla_{w} f(\theta_t, w_t, z_{j,i}) + v_j$ to server. 
	    \ENDFOR
	    \STATE Central server aggregates noisy stochastic gradients and updates the global model: $\tth_{t+1} \gets \tth_t - \eta_{\tth}\left(\frac{1}{N}\sum_{j=1}^N H_{j, \theta}\right)$ and 
	    $w_{t+1} \gets \Pi_{\WW}\left[w_t + \eta_w\left(\frac{1}{N}\sum_{j=1}^N H_{j, \theta} \right) \right]$.
		\ENDFOR
		\STATE Draw $\hat{\tth}_T$ uniformly at random from $\{\tth_t\}_{t=1}^T$. 
		\STATE \textbf{Return:} $\hat{\tth}_T$. 
	\end{algorithmic}
\end{algorithm}

\begin{assumption}
\label{ass: smooth}
\begin{enumerate}
    \item $f(\cdot, w; z)$ is $\lt$-Lipschitz and $\bt$-smooth for all $w \in \WW, z \in \ZZ$. 
    \item $f(\theta, \cdot; z)$ is $\lw$-Lipschitz, $\bw$-smooth, and $\mu$-strongly concave on $\WW$ for all $\theta \in \rdt, ~z \in \ZZ$. 
    \item $\|\naw f(\tth, w; z) - \naw f(\tth', w; z)\| \leq \btw \|\tth - \tth'\|$ and $\|\nt f(\tth, w; z) - \nt f(\tth, w'; z)\| \leq \btw \| w - w'\|$ for all $\tth, \tth', w, w', z$. 
    \item $\WW$ has $\ell_2$ diameter bounded by $D \geq 0$. 
    \item $\naw F(\theta, \ws(\theta)) = 0$ for all $\theta$, where $\ws(\theta) := \argmax_{w}F(\theta, w)$ is the unconstrained global maximizer. 
\end{enumerate}
\end{assumption}
We denote $\kw := \frac{\bw}{\mu}$ and $\ktw := \frac{\btw}{\mu}$. Also, let \[
\Phi(\theta) := \max_{w \in \WW} F(\theta, w).
\]

We can now provide our general, precise  privacy and utility guarantee for~\cref{alg: noisy Fed SGDA}:
\begin{theorem}[Privacy and Utility of~\cref{alg: noisy Fed SGDA}]
\label{thm: noisy fed SGDA convergence}
Let $\varepsilon \leq 2\ln(1/\delta), ~\delta \in (0, 1)$. Grant~Assumption~\ref{ass: smooth}. Choose $\sigma_w^2 = \frac{8 T \lw^2 \ln(1/\delta)}{\varepsilon^2 \tn^2}$, $\sigma_\tth^2 = \frac{8 T \lt^2 \ln(1/\delta)}{\varepsilon^2 \tn^2}$, and $T \geq \left(\tn \frac{\sqrt{\varepsilon}}{2 m}\right)^2$. Then~\cref{alg: noisy Fed SGDA} is $(\varepsilon, \delta)$-DP. Further, if we choose $\eta_\tth = \frac{1}{16 \kw(\bt + \btw \ktw)}$, $\eta_w = \frac{1}{\bw}$, and $T \approx \sqrt{\kw[\dph(\bt + \btw \ktw) + \btw^2 D^2]}\varepsilon \tn \sqrt{N} \min\left(\frac{1}{\lt \sqrt{\dt}}, \frac{\bw}{\btw \lw \sqrt{\kw \dw}}\right)$, then 
\begin{align*}
\expec\| \nabla \Phi(\hat{\theta}_T)\|^2 &\lesssim \sqrt{\Delta_\Phi\left(\bt + \btw \ktw)\kw + \kw \btw^2 D^2 \right)}\left[\frac{\lt \sqrt{\dt \ln(1/\delta)}}{\varepsilon \tn \sqrt{N}} + \left(\frac{\btw \sqrt{\kw}}{\bw}\right) \frac{\lw \sqrt{\dw \ln(1/\delta)}}{\varepsilon \tn \sqrt{N}}\right] \\
&\;\;\;+ \frac{\oneprime}{m N}\left(\lt^2 + \frac{\kw \btw^2 \lw^2}{\bw^2}\right).
\end{align*}
In particular, if $m \geq \min\left(\frac{\varepsilon \tn \lt}{\sqrt{N \dt \kw [\dph(\bt + \btw \ktw) + \btw^2 D^2]}}, \frac{\varepsilon \tn \lw \sqrt{\kw}}{\sqrt{N}\btw 
\bw \sqrt{\dw \kw [\dph(\bt + \btw \ktw) + \btw^2 D^2]}} \right)$, then
\[
\expec\| \nabla \Phi(\hat{\tth}_T)\|^2 \lesssim \sqrt{\kw[\dph(\bt + \btw \ktw) + \btw^2 D^2]} \left(\frac{\sqrt{\ln(1/\delta)}}{\varepsilon \tn \sqrt{N}}\right)\left(\lt \sqrt{\dt} + \left(\frac{\btw \sqrt{\kw}}{\bw}\right) \lw \sqrt{\dw}\right). 
\]
\end{theorem}

The proof of~\cref{thm: noisy fed SGDA convergence} will require several technical lemmas. These technical lemmas, in turn, require some preliminary lemmas, which we present below.

\begin{lemma}[\citet{lgr23privatefair,lin2020gradient}]
\label{lem: danskin}
Grant~Assumption~\ref{ass: smooth}.  Then $\Phi$ is $2(\bt + \btw \ktw)$-smooth with $\nabla \Phi(\tth) = \nt F(\tth, \ws(\tth))$, and $\ws(\cdot)$ is $\kw$-Lipschitz. 
\end{lemma}

\begin{lemma}[\citet{lei17}]
\label{lem: lei}
Let $\{a_l\}_{l \in [n]}$ be an arbitrary collection of vectors such that $\sum_{l=1}^{n} a_l = 0$. Further, let $\mathcal{S}$ be a uniformly random subset of $[n]$ of size $m$. Then,\[
\mathbb{E}\left\|\frac{1}{m} \sum_{l \in \mathcal{S}} a_l \right\|^2 = \frac{n - m}{(n - 1) m} \frac{1}{n}\sum_{l=1}^{n} \|a_l\|^2 \leq \frac{\mathbbm{1}_{\{m < n\}}}{m~n}\sum_{l=1}^{n}\|a_l\|^2.
\]
\end{lemma}

\begin{lemma}[Co-coercivity of the gradient]
\label{lem: coco}
For any $\beta$-smooth and convex function $g$, we have \[
\| \nabla g(a) - \nabla g(b) \|^2 \leq 2\beta (g(a) - g(b) - \langle g(b), a-b \rangle),
\]
for all $a, b \in \text{domain}(g)$.
\end{lemma}

\begin{lemma}
\label{lem: C3}
For all $t \in [T]$, the iterates of~\cref{alg: noisy Fed SGDA} satisfy \begin{align*}
\expec \Phi (\tth_t) &\leq \Phi(\theta_{t-1}) - \left(\frac{\ett}{2} - 2(\bt + \btw \ktw)\ett^2 \right)\|\nabla \Phi(\theta_{t-1})\|^2 \\
    & + \left(\frac{\ett}{2} + 2(\bt + \btw \ktw)\ett^2\right)\|\nabla \Phi(\theta_{t-1}) - \nt F(\theta_{t-1}, w_{t-1}) \|^2  + (\bt + \btw \ktw)\ett^2 \left(\dt \frac{\sigma_\theta^2}{N} + \frac{4\lt^2}{m}\mathbbm{1}_{\{ m < \tilde{n}\}}\right),
\end{align*}
conditional on $\theta_{t-1}, w_{t-1}$. 
\end{lemma}

\begin{proof}
Let us denote $\tg := \frac{1}{N} \sum_{i = 1}^{N}\left(\frac{1}{m}\sum_{i=1}^m \nt f(\tth_{t-1}, w_{t-1}; z_{ji}) + u_{t-1, i}\right) := g + u_{t-1}$, so $\tth_t = \tth_{t-1} - \eta_{\tth} \tg$, where $u_{t - 1} = \frac{1}{N}\sum_{i=1}^N u_{t-1, i}$. Since $u_{t-1, 1}, u_{t-1, 2}, ..., u_{t-1, N}$ are i.i.d. sampled from $\mathcal{N}(0, \sigma_\theta^2)$, we have that $u_{t-1} \sim \mathcal{N}(0, \frac{\sigma_\theta^2}{N})$. We proceed with the proof, conditioning on the randomness due to sampling and Gaussian noise addition.
By smoothness of $\Phi$ (see~\cref{lem: danskin}), we have 
\begin{align*}
    \Phi(\tth_t) &\leq \Phi(\tth_{t-1}) - \eta_\tth \langle \tg, \nabla \Phi(\tth_{t-1}) \rangle + (\bt + \btw \ktw) \eta_{\tth}^2 \| \tg\|^2 \\
    &= \Phi(\tth_{t-1}) - \eta_\tth \| \nabla \Phi(\tth_{t-1})\|^2 - \eta_\tth \langle \tg - \nabla \Phi(\tth_{t-1}), \nabla \Phi(\tth_{t-1}) \rangle + (\bt + \btw \ktw)\eta_\tth^2 \|\tg\|^2.
\end{align*}
Taking expectation (conditional on $\theta_{t-1}, w_{t-1}$) and using the fact that the Gaussian noise has mean zero and is independent of $(\theta_{t-1}, w_{t-1}, Z)$, plus the fact that $\expec [g] = \nt F(\theta_{t-1}, w_{t-1})$, we get
\begin{align*}
    \expec[\Phi(\theta_t)] &\leq \Phi(\theta_{t-1}) - \eta_\theta \|\nabla \Phi(\theta_{t-1})\|^2 - \ett \langle \nt F(\tth_{t-1}, w_{t-1}) - \nabla \Phi(\theta_{t-1}), \nabla \Phi(\theta_{t-1}) \rangle \\
    &\;\;\; + (\bt + \btw \ktw) \ett^2\left[\dt \frac{\sigma_\theta^2}{N} + \underbrace{\expec\|g - \nt F(\theta_{t-1}, w_{t-1})\|^2}_{\text{\textcircled{1}}} + \|\nabla_\theta F(\theta_{t-1}, w_{t-1})\|^2\right]
\end{align*}

Expanding $\expec\|g - \nt F(\theta_{t-1}, w_{t-1})\|^2$ and using Young's inequality, we get that
\begin{align*}
    \text{\textcircled{1}} & = \expec\twonorm{\frac{1}{N} \sum_{j = 1}^{N}\left(\frac{1}{m}\sum_{i=1}^m \nt f(\tth_{t-1}, w_{t-1}; z_{j,i})\right) - \frac{1}{N} \sum_{j = 1}^{N}\left(\frac{1}{\tilde{n}} \sum_{k=1}^{\tilde{n}} \nt f(\tth_{t-1}, w_{t-1}; z_{j,k})\right)}^2\\
    & = \frac{1}{N^2}\expec\twonorm{\sum_{j = 1}^{N}\frac{1}{m}\sum_{i=1}^m \left( \nt f(\tth_{t-1}, w_{t-1}; z_{j,i}) - \frac{1}{\tilde{n}} \sum_{k=1}^{\tilde{n}} \nt f(\tth_{t-1}, w_{t-1}; z_{j,k})\right)}^2\\
    &\leq \frac{1}{N}\sum_{j = 1}^{N}\expec\underbrace{\twonorm{\frac{1}{m}\sum_{i=1}^m \left( \nt f(\tth_{t-1}, w_{t-1}; z_{j,i}) - \frac{1}{\tilde{n}} \sum_{k=1}^{\tilde{n}} \nt f(\tth_{t-1}, w_{t-1}; z_{j,k})\right)}^2}_{\text{\textcircled{2}}}.
\end{align*}
For any silo $j$ , we get that
\begin{align*}
    \text{\textcircled{2}} &\leq \frac{\mathbbm{1}_{\{m < \tilde{n}\}}}{m~\tilde{n}}\sum_{l=1}^{\tilde{n}}\twonorm{\nt f(\tth_{t-1}, w_{t-1}; z_{j,l}) - \frac{1}{\tilde{n}} \sum_{i=1}^{\tilde{n}} \nt f(\tth_{t-1}, w_{t-1}; z_{j,i})}^2 \\
    &\leq \frac{\mathbbm{1}_{\{m < \tilde{n}\}}}{m~\tilde{n}}\sum_{l=1}^{\tilde{n}}\twonorm{\frac{1}{\tilde{n}}\sum_{i=1}^{\tilde{n}}\nt f(\tth_{t-1}, w_{t-1}; z_{j,l}) - \frac{1}{\tilde{n}} \sum_{i=1}^{\tilde{n}} \nt f(\tth_{t-1}, w_{t-1}; z_{j,i})}^2 \\
    &\leq \frac{\mathbbm{1}_{\{m < \tilde{n}\}}}{m~\tilde{n}^2} \sum_{l=1}^{\tilde{n}}\sum_{i=1}^{\tilde{n}}\twonorm{\nt f(\tth_{t-1}, w_{t-1}; z_{j,l}) -  \nt f(\tth_{t-1}, w_{t-1}; z_{j,i})}^2 \leq \frac{4\mathbbm{1}_{\{m < \tilde{n}\}}L_\theta^2}{m}.
\end{align*}

We used \cref{lem: lei} in the first inequality. In the second inequality, we used Young's inequality and Lipschitz continuity of $f$. The remainder of the proof follows from \cite{lgr23privatefair}. We restate the further steps for ease. Thus, we get that
\begin{align*}
    \expec[\Phi(\theta_t)] &\leq \Phi(\theta_{t-1}) - \eta_\theta \|\nabla \Phi(\theta_{t-1})\|^2 - \ett \langle \nt F(\tth_{t-1}, w_{t-1}) - \nabla \Phi(\theta_{t-1}), \nabla \Phi(\theta_{t-1}) \rangle \\
    &\;\;\;+ (\bt + \btw \ktw) \ett^2\left[\dt \frac{\sigma_\theta^2}{N} + \frac{4\lt^2}{m}\mathbbm{1}_{\{m<\tilde{n}\}} + \|\nabla_\theta F(\theta_{t-1}, w_{t-1})\|^2\right]\\
     &\leq \Phi(\theta_{t-1}) - \eta_\theta \|\nabla \Phi(\theta_{t-1})\|^2 - \ett \langle \nt F(\tth_{t-1}, w_{t-1}) - \nabla \Phi(\theta_{t-1}), \nabla \Phi(\theta_{t-1}) \rangle \\
    &\;\;\;+ (\bt + \btw \ktw) \ett^2\left[\dt \frac{\sigma_\theta^2}{N} + \frac{4\lt^2}{m}\mathbbm{1}_{\{m<\tilde{n}\}} + 2\|\nabla_\theta F(\theta_{t-1}, w_{t-1}) - \nabla \Phi(\theta_{t-1})\|^2 + 2\|\nabla \Phi(\theta_{t-1}) \|^2\right] \\
     &\leq \Phi(\theta_{t-1}) - \eta_\theta \|\nabla \Phi(\theta_{t-1})\|^2 + \frac{\ett}{2}\left[\| \nabla \Phi(\theta_{t-1}) - \nt F(\theta_{t-1}, w_{t-1})\|^2 + \| \nabla \Phi(\theta_{t-1}) \|^2\right] \\
    &\;\;\;+ (\bt + \btw \ktw) \ett^2\left[\dt \frac{\sigma_\theta^2}{N} + \frac{4\lt^2}{m}\mathbbm{1}_{\{m<\tilde{n}\}} + 2\|\nabla_\theta F(\theta_{t-1}, w_{t-1}) - \nabla \Phi(\theta_{t-1})\|^2 + 2\|\nabla \Phi(\theta_{t-1})\|^2\right] \\
    &\leq \Phi(\theta_{t-1}) - \left(\frac{\ett}{2} - 2(\bt + \btw \ktw)\ett^2 \right)\|\nabla \Phi(\theta_{t-1})\|^2 \\
    &\;\;\; + \left(\frac{\ett}{2} + 2(\bt + \btw \ktw)\ett^2\right)\|\nabla \Phi(\theta_{t-1}) - \nt F(\theta_{t-1}, w_{t-1}) \|^2  \\
    &\;\;\;+ (\bt + \btw \ktw)\ett^2 \left(\dt \frac{\sigma_\theta^2}{N} + \frac{4\lt^2}{m}\mathbbm{1}_{\{ m < \tilde{n}\}}\right).
\end{align*}
In the second and third inequalities, we used Young's inequality and Cauchy-Schwartz.  
\end{proof}

\begin{lemma}
\label{lem: C5}
Grant Assumption~\ref{ass: smooth}. If $\eta_\tth = \frac{1}{16 \kw(\bt + \btw \ktw)}$, then the iterates of~\cref{alg: noisy Fed SGDA} satisfy ($\forall t \geq 0$)\[
\expec \Phi(\theta_{t+1}) \leq \expec\left[\Phi(\theta_t) - \frac{3}{8}\ett \|\Phi(\theta_t)\|^2 + \frac{5}{8}\ett\left(\btw^2 \|\ws(\theta_t) - w_t\|^2 + \dt \frac{\sigma_\theta^2}{N} + \frac{4\lt^2}{m}\oneprime \right)\right].
\]
\end{lemma}
With some changes to the noise variance, the proof follows exactly from \cite{lgr23privatefair}. We restate it here for ease.
\begin{proof}
By~\cref{lem: C3}, we have \begin{align*}
\expec \Phi (\tth_{t+1}) &\leq \expec \Phi (\tth_{t}) - \left(\frac{\eta_{\tth}}{2} - 2(\bt + \btw \ktw)\eta_{\tth}^2 \right)\expec\| \nabla \Phi (\tth_{t})\|^2 \\
&\;\;\; + \left(\frac{\eta_{\tth}}{2} + 2\eta_\tth^2(\bt + \btw \ktw)\expec\|\nabla \Phi(\tth_{t}) - \nt F(\tth_{t}, w_{t})\|^2 \right) + (\bt + \btw \ktw) \eta_{\tth}^2\left(\dt \frac{\sigma_\tth^2}{N} + \frac{4\lt^2}{m} \mathbbm{1}_{\{m < \tilde{n}\}}\right) \\
&\leq \expec \Phi(\theta_t) - \frac{3}{8}\eta_\tth \expec\|\nabla \Phi(\theta_t) \|^2 + \frac{5}{8} \eta_\tth \left[\expec\|\nabla \Phi(\theta_t) - \nt F(\theta_t, w_t)\|^2 + \dt \frac{\sigma_\tth^2}{N} + \frac{4\lt^2}{m}\oneprime \right] \\
& \leq \expec \Phi(\theta_t) - \frac{3}{8}\eta_\tth \expec\|\nabla \Phi(\theta_t) \|^2 + \frac{5}{8} \eta_\tth \left[\btw^2 \expec\|\ws(\theta_t) - w_t\|^2 + \dt \frac{\sigma_\tth^2}{N} + \frac{4\lt^2}{m}\oneprime \right].
\end{align*}
In the second inequality, we simply used the definition of $\ett$. In the third inequality, we used the fact that $\nabla \Phi(\theta_t) = \nt F(\theta_t, \ws(\theta_t))$ (see~\cref{lem: danskin}) together with~Assumption~\ref{ass: smooth} (part 3).
\end{proof}

\begin{lemma}
\label{lem: C4}
Grant Assumption~\ref{ass: smooth}. If $\etw = \frac{1}{\bw}$, then the iterates of~\cref{alg: noisy Fed SGDA} satisfy ($\forall t \geq 0$) \begin{align*}
    \expec\|\ws(\theta_{t+1}) - w_{t+1}\|^2 &\leq \left(1 - \frac{1}{2\kw} + 4\kw \ktw^2 \ett^2 \btw^2 \right)\expec\|\ws(\theta_{t}) - w_{t}\|^2 + \frac{2}{\bw^2}\left(\frac{4\lw^2}{m}\mathbbm{1}_{\{m < \tilde{n}\}} + \dw \frac{\sigma_w^2}{N}\right) \\
    &\;\;\; + 4\kw \ktw^2 \ett^2 \left(\expec\|\nabla \Phi(\theta_t)\|^2 + \dt \frac{\sigma_{\theta}^2}{N} 
    \right).
\end{align*}
\end{lemma}
\begin{proof}

Fix any t and denote $\tg_w := \frac{1}{N} \sum_{i = 1}^{N}\left(\frac{1}{m}\sum_{i=1}^m \naw f(\theta_{t-1}, w_{t-1}; z_{ji}) + v_{t-1, i}\right) := g_w + v_{t-1}$, so $w_t = w_{t-1} - \eta_{\tth} \tg_w$, where $v_{t - 1} = \frac{1}{N}\sum_{i=1}^N v_{t-1, i}$. Since $v_{t-1, 1}, v_{t-1, 2}, ..., v_{t-1, N}$ are i.i.d. sampled from $\mathcal{N}(0, \sigma_w^2)$, we have that $v_{t-1} \sim \mathcal{N}(0, \frac{\sigma_w^2}{N})$. Now take $\delta_t := \expec\|\ws(\theta_t) - w_t\|^2 := \expec\|\ws - w_t\|^2$. We may assume without loss of generality that $f(\theta, \cdot; z)$ is $\mu$-strongly \textit{convex} and that $w_{t+1} = \Pi_{\WW}[w_t - \frac{1}{\bw}\widetilde{g_w}] := \Pi_{\WW}[w_t - \frac{1}{\bw}\left(g_w + v_t\right)]$.
Now, \begin{align*}
  \expec\|w_{t+1} - \ws\|^2 &= \expec\left\|\Pi_{\WW}[w_t - \frac{1}{\bw}\widetilde{g_w}]- \ws\right\|^2 \leq \expec\left\|w_t - \frac{1}{\bw}\widetilde{g_w} - \ws\right\|^2 \\
  &=\expec\|w_t - \ws\|^2 + \frac{1}{\bw^2}\left[\expec\|g_w\|^2 + \dw \frac{\sigma_w^2}{N} \right] - \frac{2}{\bw} \expec\left\langle w_t - \ws, \widetilde{g_w} \right\rangle \\
  &\leq \expec\|w_t - \ws\|^2 + \frac{1}{\bw^2}\left[\expec\|g_w\|^2 + \dw \frac{\sigma_w^2}{N} \right] - \frac{2}{\bw} \expec\left[F(\theta_t, w_t) - F(\theta_t, \ws) + \frac{\mu}{2}\|w_t - \ws\|^2\right] \\
  &\leq \delta_t\left(1 - \frac{\mu}{\bw} \right) - \frac{2}{\bw}\expec\left[F(\theta_t, w_t) - F(\theta_t, \ws) \right] + \frac{\expec\|g_w\|^2}{\bw^2} + \frac{\dw \sigma_w^2}{N\bw^2}.
\end{align*} 

Rewriting, \begin{align*}
    \expec\| g_w \|^2 &= \expec\left[\| g_w - \naw
    F(\theta_t, w_t)\|^2 + \|\naw F(\theta_t, w_t)\|^2 \right]\\
    &\leq \expec\left[\| g_w - \naw
    F(\theta_t, w_t)\|^2 \right] + 2\bw[F(\theta_t, w_t) - F(\theta_t, \ws(\theta_t))],
\end{align*}
using $\expec g_w = \naw
    F(\theta_t, w_t)$ in the first equality, and~\cref{lem: coco} (plus~Assumption~\ref{ass: smooth} part 5) in the second inequality.

Expanding, $\expec\left[\|g_w - \naw F(\theta_t, w_t)\|^2 \right]$ in the same manner as we did for $\expec\left[\|g - \nabla_\theta F(\theta_t, w_t)\|^2 \right]$ in \cref{lem: C3}, we get that
\begin{align*}
    \expec\left[\|g_w - \naw
    F(\theta_t, w_t)\|^2 \right] \leq \frac{4\mathbbm{1}_{\{m < \tilde{n}\}}L_w^2}{m}
\end{align*}

This implies \begin{align}
\label{eq: star}
    \expec\|w_{t+1} - \ws\|^2 &\leq \delta_t\left(1 - \frac{1}{\kw}\right) + \frac{1}{\bw^2}\left[\dw \frac{\sigma_w^2}{N} + \frac{4\lw^2}{m}\oneprime \right].
\end{align}

Therefore, \begin{align*}
\delta_{t+1} &= \expec\|w_{t+1} - \ws(\theta_t) + \ws(\theta_t) - \ws(\theta_{t+1})\|^2\\
&\leq \left(1 + \frac{1}{2\kw - 1}\right)\expec\|w_{t+1} - \ws(\theta_t)\|^2 + 2\kw \expec\|\ws(\theta_t) - \ws(\theta_{t+1})\|^2 \\
&\leq \left(1 + \frac{1}{2\kw - 1}\right)\left(1 - \frac{1}{\kw}\right)\delta_t + \frac{2}{\bw^2}\left[\dw \frac{\sigma_w^2}{N} + \frac{4\lw^2}{m}\oneprime \right] + 2\kw \expec\|\ws(\theta_t) - \ws(\theta_{t+1})\|^2 \\
&\leq \left(1 + \frac{1}{2\kw - 1}\right)\left(1 - \frac{1}{\kw}\right)\delta_t + \frac{2}{\bw^2}\left[\dw \frac{\sigma_w^2}{N} + \frac{4\lw^2}{m}\oneprime \right] + 2\kw \ktw^2 \expec \|\theta_t - \theta_{t+1}\|^2 \\
&\leq \left(1 + \frac{1}{2\kw - 1}\right)\left(1 - \frac{1}{\kw}\right)\delta_t + \frac{2}{\bw^2}\left[\dw \frac{\sigma_w^2}{N} + \frac{4\lw^2}{m}\oneprime \right] \\
&\;\;\;+ 4\kw \ktw^2 \ett^2\left[\expec\|\nt F(\theta_t, w_t) - \nabla \Phi(\theta_t) \|^2 + \|\nabla \Phi(\theta_t)\|^2 + \dt \frac{\sigma_\theta^2}{N} \right] \\
&= \left(1 + \frac{1}{2\kw - 1}\right)\left(1 - \frac{1}{\kw}\right)\delta_t + \frac{2}{\bw^2}\left[\dw \frac{\sigma_w^2}{N} + \frac{4\lw^2}{m}\oneprime \right] \\
&\;\;\;+ 4\kw \ktw^2 \ett^2\left[\expec\|\nt F(\theta_t, w_t) - \nt F(\theta_t, \ws(\theta_t) \|^2 + \|\nabla \Phi(\theta_t)\|^2 + \dt \frac{\sigma_\theta^2}{N} \right]\\
&\leq \left(1 + \frac{1}{2\kw - 1}\right)\left(1 - \frac{1}{\kw}\right)\delta_t + \frac{2}{\bw^2}\left[\dw \frac{\sigma_w^2}{N} + \frac{4\lw^2}{m}\oneprime \right] \\
&\;\;\;+ 4\kw \ktw^2 \ett^2\left[\btw^2\expec\|w_t - \ws(\theta_t)\|^2 + \|\nabla \Phi(\theta_t)\|^2 + \dt \frac{\sigma_\theta^2}{N} \right],
\end{align*}
by Young's inequality, \cref{eq: star}, and \cref{lem: danskin}. Since $\left(1 + \frac{1}{2\kw - 1}\right)\left(1 - \frac{1}{\kw}\right) \leq 1 - \frac{1}{2\kw}$, we obtain \begin{align*}
\delta_{t+1} &\leq \left(1 - \frac{1}{2\kw} +  4\kw \ktw^2 \ett^2 \btw^2 \right)\delta_t + \frac{2}{\bw^2}\left[\dw \frac{\sigma_w^2}{N} + \frac{4\lw^2}{m}\oneprime \right] + 4\kw \ktw^2 \ett^2\left[\|\nabla \Phi(\theta_t)\|^2 + \dt \frac{\sigma_\theta^2}{N} \right], 
\end{align*}
as desired. 
\end{proof}

\begin{proof} (of~\cref{thm: noisy fed SGDA convergence})
\textbf{Privacy:} By the definition of ISRL-DP, independence of the Gaussian noise across silos, and symmetry of~\cref{alg: noisy Fed SGDA} w.r.t. every silo, it suffices to show that the full transcript of silo $j$'s communications is $(\varepsilon, \delta)$-DP for any fixed settings of other silos' messages and data. But, the prescribed choices of noise in~\cref{thm: noisy fed SGDA convergence} follows directly from the calculations of Theorem E.1 \cite{lgr23privatefair} \textit{with respect to a single silo}. Hence, the privacy of the communication transcript of data  follows directly from the analysis of \cite{lgr23privatefair}. Therefore, \cref{alg: noisy Fed SGDA} is $(\varepsilon, \delta)$-ISRL-DP. 
 
\textbf{Convergence:}

Denote $\zeta := 1 - \frac{1}{2\kw} + 4 \kw \ktw^2 \ett^2 \btw^2$, ~$\delta_t = \expec\|\ws(\theta_t) - w_t\|^2$, and 
\[
C_t := \frac{2}{\bw^2}\left(\frac{4\lw^2}{m}\mathbbm{1}_{\{m < \tilde{n}\}} + \dw \frac{\sigma_w^2}{N}\right) + 4\kw \ktw^2 \ett^2 \left(\expec\|\nabla \Phi(\theta_t)\|^2 + \dt \frac{\sigma_{\theta}^2}{N}\right),\] 
so that~\cref{lem: C4} reads as \begin{equation}
\label{eq: C4}
\delta_{t} \leq \zeta \delta_{t-1} + C_{t-1} 
\end{equation}
for all $t \in [T]$. 
Applying~\cref{eq: C4} recursively, we have \begin{align*}
    \delta_t &\leq \zeta^t \delta_0 + \sum_{j=0}^{t-1} C_{t-j-1} \zeta^j \\
    &\leq \zeta^t D^2 + 4\kw \ktw^2 \eta_\tth^2 \sum_{j=0}^{t-1} \zeta^{t-1-j} \expec\|\nabla \Phi(\theta_j)\|^2 \\
    &\;\;\;+\left(\sum_{j=0}^{t-1} \zeta^{t-1-j} \right)\left[\frac{2}{\bw^2}\left(\frac{4 \lw^2}{m}\oneprime + \dw \frac{\sigma_w^2}{N} \right) + 4 \kw \ktw^2 \ett^2 \dt \frac{\sigma_{\theta}^2}{N} \right].
\end{align*} 
Combining this inequality with~\cref{lem: C5}, we get \begin{align*}
    \expec\Phi(\theta_t) &\leq \expec\left[\Phi(\theta_{t-1}) - \frac{3}{8} \ett \|\nabla \Phi(\theta_{t-1}) \|^2 \right] + \frac{5}{8} \ett\left(\dt \frac{\sigma_\theta^2}{N} + \frac{4 \lt^2}{m} \oneprime \right) \\
    &\;\;\;+ \frac{5}{8} \ett \btw^2\Bigg\{\zeta^t D^2 + 4\kw \ktw^2 \ett^2 \sum_{j=0}^{t-1} \zeta^{t-1-j} \expec\| \nabla \Phi(\theta_j)\|^2 \\
    &\;\;\;\;\;\;+ \left(\sum_{j=0}^{t-1} \zeta^{t-1-j} \right)\left[\frac{2}{\bw^2}\left(\frac{4 \lw^2}{m}\oneprime + \dw \frac{\sigma_w^2}{N} \right) + 4\kw\ktw^2 \ett^2 \dt \frac{\sigma_\theta^2}{N} \right] \Bigg\}.
\end{align*}
Summing over all $t \in [T]$ and re-arranging terms yields \begin{align*}
\expec \Phi(\theta_T) &\leq \Phi(\theta_0) - \frac{3}{8} \ett \sum_{t=1}^T \expec\|\nabla \Phi(\theta_{t-1}) \|^2 + \frac{5}{8} \ett T\left(\dt \frac{\sigma_\theta^2}{N} + \frac{4 \lt^2}{m}\oneprime \right) + \frac{5}{8}\ett \btw^2 \left(\sum_{t=1}^T \zeta^t\right) D^2 \\
&\;\;\;+ 4\ett^3 \btw^2 \kw \ktw^2 \sum_{t=1}^T \sum_{j=0}^{t-1} \zeta^{t-1-j} \expec \|\nabla \Phi(\theta_j)\|^2 
\\
&\;\;\;
+ \frac{5}{8}\left(\sum_{t=1}^T \sum_{j=0}^{t-1} \zeta^{t-1-j} \right)\ett \btw^2 \left[\frac{2}{\bw^2}\left(\frac{4\lw^2}{m}\oneprime + \dw \frac{\sigma^2_w}{N} \right) + 4\kw \ktw^2 \ett^2 \dt \frac{\sigma_\theta^2}{N} \right].
\end{align*}
Now, $\zeta \leq 1 - \frac{1}{4 \kw}$, which implies that \begin{align*}
\sum_{t=1}^T \zeta^t \leq 4\kw~~\text{and} \\
\sum_{t=1}^T \sum_{j=0}^{t-1} \zeta^{t-1-j} \leq 4 \kw T.
\end{align*}
Hence \begin{align*}
\frac{1}{T} \sum_{t=1}^T \expec\|\nabla \Phi(\theta_t)\|^2 &\leq \frac{3[\Phi(\theta_0) - \expec \Phi(\theta_T)]}{\ett T} + \frac{5}{3}\left(\dt \frac{\sigma_\theta^2}{N} + \frac{4\lt^2}{m}\oneprime \right) +  \frac{7 \btw^2 D^2 \kw}{T} \\
&\;\;\;+ \frac{48 \ett^2 \btw^2 \kw^2 \ktw^2}{T}\left(\sum_{t=1}^T \expec\|\nabla \Phi(\theta_t) \|^2 \right) \\
&\;\;\;+ 8\kw \btw^2
\frac{2}{\bw^2}\left(\frac{4\lw^2}{m}\oneprime + \dw \frac{\sigma_w^2}{N} \right) + 32 \btw^2 \kw^2 \ktw^2 \ett^2 \dt \frac{\sigma_\theta^2}{N} 
.
\end{align*}
Since $\ett^2 \btw^2 \kw^2 \ktw^2 \leq \frac{1}{256}$, we obtain \begin{align*}
\expec\| \nabla \Phi(\hat{\theta}_T)\|^2 &\lesssim \frac{\Delta_\Phi \kw}{T}\left(\bt + \btw \ktw \right) + \frac{\dt \lt^2 T \ln(1/\delta)}{\varepsilon^2 \tn^2 N} + \frac{1}{m}\oneprime\left(\lt^2 + \frac{\kw \btw^2 \lw^2}{\bw^2} \right) + \frac{\kw \btw^2 \lw^2 \dw T \ln(1/\delta)}{\bw^2 \varepsilon^2 \tn^2 N} \\
&\;\;\;+ \frac{\btw^2 D^2 \kw}{T}.
\end{align*}
Our choice of $T$ then implies \begin{align*}
\expec\| \nabla \Phi(\hat{\theta}_T)\|^2 &\lesssim \sqrt{\Delta_\Phi\left(\bt + \btw \ktw)\kw + \kw \btw^2 D^2 \right)}\left[\frac{\lt \sqrt{\dt \ln(1/\delta)}}{\varepsilon \tn \sqrt{N}} + \left(\frac{\btw \sqrt{\kw}}{\bw}\right) \frac{\lw \sqrt{\dw \ln(1/\delta)}}{\varepsilon \tn \sqrt{N}}\right] \\
&\;\;\;+ \frac{\one}{m}\left(\lt^2 + \frac{\kw \btw^2 \lw^2}{\bw^2}\right).
\end{align*}
Finally, our choice of sufficiently large $m$ yields the last claim in~\cref{thm: noisy fed SGDA convergence}. 
\end{proof}

\subsection{Proof of Theorem~\ref{thm: steffle utility}: Utility Claim 
}
The utility claim in~\cref{thm: steffle utility} is an easy consequence of~\cref{thm: noisy fed SGDA convergence}, which we proved above: 
\begin{theorem}[Precise Re-statement of Utility Claim in~\cref{thm: steffle utility}]

\label{thm: fed FERMI utility}
Assume the loss function $\ell(\cdot, x, y)$ and $\FF(x, \cdot)$ are Lipschitz continuous with Lipschitz gradient for all $(x, y)$, and $\widehat{P}_S(r) \geq \rho > 0 ~\forall~r \in [k]$. In~\cref{alg: steffle}, choose $\WW$ to be a sufficiently large ball that contains $W^*(\theta) := \argmax_{W} \widehat{F}(\theta, W)$ for every $\theta$ in some neighborhood of $\theta^* \in \argmin_{\theta} \max_{W} \widehat{F}(\theta, W)$. Then there exist algorithmic parameters such that the $(\varepsilon, \delta)$-ISRL-DP~\cref{alg: steffle} returns $\hat{\theta}_T$ with \[
\expec\|\nabla \text{FERMI}(\hat{\theta}_T)\|^2 = \mathcal{O}\left(\frac{\sqrt{\max(d_{\theta}, kl) \ln(1/\delta)}}{\varepsilon \tn \sqrt{N}}\right),
\]
treating $D = \text{diameter}(\WW)$, $\lambda$, $\rho$, and the Lipschitz and smoothness parameters of $\ell$ and $\FF$ as constants.
\end{theorem}
\begin{proof}
By~\cref{thm: noisy fed SGDA convergence}, it suffices to show that Assumption~\ref{ass: smooth} holds for $f(\theta, W; z_i) := \ell(\theta, x_i, y_i) + \lambda \widehat{\psi}_i(\theta, W)$. 
We assumed $\ell(\cdot, x_i, y_i)$ is Lipschitz continuous with Lipschitz gradient. Further, the work of~\citet{fermi} showed that $f(\theta, \cdot; z_i)$ is strongly concave. Thus, it suffices to show that $\widehat{\psi}_i(\theta, W)$ is Lipschitz continuous with Lipschitz gradient.
\end{proof}

\section{Complete Version of Theorem~\ref{thm: informal Fermi as minmax}}
\label{app: formal minmax}

Let $\widehat{\mathbf{y}}(x_{ji}; \theta)   \in \{0,1\}^l$ and $\mathbf{s}_{ji} \in\{0,1\}^k$ be the one-hot encodings of $\widehat{y}(x_{ji}, \theta)$ and $s_{ji}$, respectively: i.e., $\widehat{\mathbf{y}}_h(x_i; \theta) = \mathbbm{1}_{\{\widehat{y}(x_{ji}, \theta) = h\}}$ and $\mathbf{s}_{{ji},r} = \mathbbm{1}_{\{s_{ji} = r\}}$  for $h \in [l], r \in [k]$. Also, denote $\widehat{P}_{s} = {\rm diag}({\widehat{p}}_{S}(1), \ldots, \widehat{p}_{S}(k))$, where $\widehat{p}_{S}(r) := \frac{1}{N\tilde{n}}\sum_{j=1}^N\sum_{i=1}^{\tilde{n}} \mathbbm{1}_{\{s_{ji} = r \}} \geq \rho > 0$ is the empirical probability of attribute $r$ ($r \in [k]$).
Then we have the following re-formulation of~\cref{eq: FERMI} as a min-max problem:
\begin{theorem}[\citet{fermi}]
\label{thm: Fermi as minmax}
\cref{eq: FERMI} is equivalent to
\small
\begin{align}
\label{eq: empirical minmax1}
\min_{\theta} \max_{W \in \mathbb{R}^{k \times l}} 
\left\{
\widehat{F}(\theta, W) := 
\widehat{\mathcal{L}}(\theta) + \lambda \frac{1}{N \tilde{n}} \sum_{j=1}^N \sum_{i=1}^{\tilde{n}} \widehat{\psi}_{ji}(\theta, W)
\right\}.
\end{align}
\normalsize
where 
\small
\begin{align*}
  \widehat{\psi}_{ji}(\theta, W) &:=  -\Tr(W \expec[\widehat{\by}(x_{ji}, \theta) \widehat{\by}(x_{ji}, \theta)^T | x_i] 
  W^T) 
  \\
  &\;\;\;\; + 2 \Tr(W \expec[\widehat{\by}(x_{ji}; \theta) \bs_{ji}^T | x_{ji}, \bs_{ji}] \widehat{P}_{s}^{-1/2}) - 1,  
\end{align*}
\normalsize
$\expec[\widehat{\by}(x_{ji}; \theta) \widehat{\by}(x_{ji}; \theta)^T | x_{ji}] = {\rm diag}(\FF_1(x_{ji}, \theta), \ldots, \FF_l(x_{ji}, \theta))$,
and $\expec[\widehat{\by}(x_{ji}; \theta) \bs_{ji}^T | x_{ji}, \bs_{ji}]$ is a $k \times l$ matrix with $\expec[\widehat{\by}(x_{ji}; \theta) \bs_{ji}^T | x_{ji}, \bs_{ji}]_{r,u} = \bs_{{ji}, r} \FF_u(x_{ji}, \theta)$.
\end{theorem}
Strong concavity of $\widehat{\psi}_i$ is shown in~\cite{fermi}.

\section{Numerical Experiments: Additional Details and Results}
\label{app: experiments}
The code for this work can be found at \href{https://github.com/justaguyalways/Stochastic-Federated-Differentially-Private-and-Fair-Learning}{https://github.com/justaguyalways/Stochastic-Federated-Differentially-Private-and-Fair-Learning}.
\subsection{Measuring Demographic Parity and Equalized Odds Violation}
\label{ssec:eqodds_steffle_variant}
Following the evaluation setup from \cite{lgr23privatefair}. We used the expressions given in~\cref{eq:demopairviolation} and~\cref{eq:eqoddsviolation} to measure the demographic parity violation and the equalized odds violation respectively. We denote $\mathcal{Y}$ to be the set of all possible output classes and $\mathcal{S}$ to be the classes of the sensitive attribute. $P[E]$ denotes the empirical probability of the occurrence of an event E. We subsequently present the results for the Demographic Parity and misclassification error trade-offs for the Adult and Retired Adult Datasets and results for Equalized odds on the credit card dataset.

\begin{equation}
\label{eq:demopairviolation}
    \max_{y' \in \mathcal{Y}, s_1, s_2 \in \mathcal{S}} \left|P[\haty = y' | s = s_1] - P[\haty = y' | s = s_2]\right|
\end{equation}

\begin{equation}
\label{eq:eqoddsviolation}
\begin{split}
    \max_{y' \in \mathcal{Y}, s_1, s_2 \in \mathcal{S}} \max(\left|P[\haty = y' | s = s_1, y = y'] - P[\haty = y' | s = s_2, y = y']\right|, \\
    \left|P[\haty = y' | s = s_1, y \neq y'] - P[\haty = y' | s = s_2, y \neq y']\right|)
\end{split}
\end{equation}

\subsection{Tabular Datasets}
\label{app: tabular}
\subsubsection{Model Description and Experimental Details}
\noindent \textbf{Demographic Parity: }We split each dataset in a 3:1 train:test ratio. We preprocess the data similar to~\citet{hardt2016equality} and use a simple logistic regression model with a sigmoid output $O = \sigma(Wx + b)$ which we treat as conditional probabilities $p(\haty = i |x)$. The predicted variables and sensitive attributes are both binary in this case across all the datasets. We analyze fairness-accuracy trade-offs with three different values of $\varepsilon \in \{1, 3, 9\}$ for each dataset. We compare against state-of-the-art algorithms proposed in~\citet{ling2024fedfdp} and (the demographic parity objective of) \citet{tran2021differentially}. The tradeoff curves for SteFFLe were generated by sweeping across different values for $\lambda \in [0, 2.0]$. The learning rates for the descent and ascent, $\eta_\theta$ and $\eta_w$ of which the former was subjected to a step decay of 0.8 every 10 epochs while the latter was kept constant during the optimization process with the initial values at $0.25$ and $1e-5$, respectively. Batch size was 256. We tuned the $\ell_2$ diameter of the projection set $\WW$ and $\theta$-gradient clipping threshold in $[1,5]$ in order to generate stable results with high privacy (i.e. low $\varepsilon$). Each model was trained for 40 epochs. The results displayed are averages over 15 trials (random seeds) for each value of $\varepsilon$ and heterogeneity level $h$. 

\noindent \textbf{Equalized Odds:} We replicated the experimental setup described above, but we took $\ell_2$ diameter of $\WW$ and the value of gradient clipping for $\theta$ to be in $[1,2]$. Also, we only tested two values of $\varepsilon \in \{1, 3\}$ and two values of heterogeneity levels $h \in \{0, 0.75\}$.

\subsubsection{Description of Datasets}
\label{app:datasetdesc}
\noindent \textbf{Adult Income Dataset:} This dataset contains the census information about the individuals. The classification task is to predict whether the person earns more than 50k every year or not. We followed a preprocessing approach similar to \citet{fermi}. After preprocessing, there were a total of 102 input features from this dataset. The sensitive attribute for this work in this dataset was taken to be gender. This dataset consists of around 48,000 entries spanning across two CSV files, which we combine and then we take the train-test split of 3:1.

\noindent \textbf{Retired Adult Income Dataset:} The Retired Adult Income Dataset proposed by~\citet{retiredadult} is essentially a superset of the Adult Income Dataset which attempts to counter some caveats of the Adult dataset. The input and the output attributes for this dataset is the same as that of the Adult Dataset and the sensitive attribute considered in this work is the same as that of the Adult. This dataset contains around 45,000 entries.

\noindent \textbf{Credit Card Dataset:} This dataset contains the financial data of users in a bank in Taiwan consisting of their gender, education level, age, marital status, previous bills, and payments. The assigned classification task is to predict whether the person defaults their credit card bills or not, essentially making the task if the clients were credible or not. We followed a preprocessing approach similar to \citet{fermi}. After preprocessing, there were a total of 85 input features from this dataset. The sensitive attribute for this dataset was taken to be gender. This dataset consists of around 30,000 entries from which we take the train-test split of 3:1.

\subsubsection{Modelling Heterogeneity}

The proposed algorithm, Algorithm~\ref{alg : heteromodelling}, partitions a dataset $D$ of size $n$ into $K$ silos, incorporating a specified heterogeneity level $h$. It begins by dividing the data into equal-sized partitions based on a predefined attribute $A$ (age, in all our experiments), which determines initial partition labels. For each silo $k$, a subset of data points is sampled based on $h$, introducing a controlled deviation from homogeneous partitioning. This ensures that each silo receives approximately $\frac{n}{K}$ data points while allowing for varying levels of overlap across silos. The remaining unassigned points are then distributed to maintain a balance between homogeneous and heterogeneous data distributions. The final output consists of $K$ subsets $\{D_1, D_2, \dots, D_K\}$, with $h$ governing the diversity of the partitions.

This algorithm offers notable advantages over traditional methods using Dirichlet distributions for modeling heterogeneity in federated learning. In terms of \textit{interpretability}, it provides explicit control over the degree of heterogeneity via the parameter $h$, allowing for easy quantification and adjustment of deviations from homogeneity. In contrast, Dirichlet-based methods abstract this process, making it harder to control or interpret heterogeneity without post-hoc analysis.

In terms of \textit{flexibility}, the algorithm allows granular control over partitioning based on specific attributes and sample sizes, which is not straightforward with Dirichlet-based methods. While Dirichlet distributions often lead to unpredictable or uneven partition sizes, this algorithm ensures each silo receives approximately $\frac{n}{K}$ data points, with controlled deviations through $h$. This flexibility makes it a more practical and customizable solution for federated learning scenarios requiring balanced silo sizes and heterogeneous distributions.

Thus, the proposed approach improves both interpretability and flexibility, making it more effective for modeling heterogeneity in federated learning.

\begin{algorithm}[t]
    \caption{Modelling Heterogeneity Levels in Silo Data}
    \begin{algorithmic}[1]
        \STATE \textbf{Input:} Set of data points $D$, number of silos $K$, total points $n$, attribute for partitioning $A$, level of heterogeneity $h$
        \STATE \textbf{Output:} Partitioned subsets $\{D_1, D_2, \dots, D_K\}$
        
        \STATE Divide data into equal sized partitions indexed $L \in \{0, 1, \dots, K-1\}^n$ based on attribute $A$.
        \STATE Initialize vector $P$ of length $n$, representing initial partition assignments.
        
        \STATE Initialize list $S = []$, to store indices for each silo.
        
        \FOR{each silo $k$ from $0$ to $K-1$}
            \STATE Let $I_k = \{i \mid P[i] = k\}$, indices where partition $P$ is $k$.
            \STATE Sample $m_k = \left\lfloor \frac{n}{K} \times h \right\rfloor$ unique indices from $I_k$ without replacement.
            \STATE Set $P[m_k]$ to an unused label (e.g., $K$), indicating these indices are selected.
            \STATE Add $m_k$ to list $S_k$ in $S$.
        \ENDFOR
        
        \FOR{each silo $k$ from $0$ to $K-1$}
            \STATE Let $C = \{i \mid P[i] \neq K\}$, indices not yet assigned.
            \STATE Calculate remaining indices needed: $r_k = \frac{n}{K} - |S_k|$.
            \STATE Sample $r_k$ indices from $C$ without replacement.
            \STATE Update $P$ for these indices to $K$, marking them as selected.
            \STATE Add these indices to the respective list $S_k$ in $S$.
        \ENDFOR
        
        \STATE Update $P$ such that all indices in each $S_k$ are set to $k$, forming final partitions.

        \STATE Split $D$ into $\{D_1, D_2, \dots, D_K\}$ based on updated partitions $P$.
        
        \STATE \textbf{return} $\{D_1, D_2, \dots, D_K\}$
    \end{algorithmic}
    \label{alg : heteromodelling}
\end{algorithm}

\subsection{Additional Numerical Results}

\subsubsection{Empirical Observations}

The experiments with demographic parity and equalized odds (fairness) provide several key insights into the trade-off between misclassification error and fairness across all groups. Across all plots, our model consistently outperforms the baselines proposed by \cite{tran2021differentially} and \cite{ling2024fedfdp}, exhibiting substantial performance improvements. Furthermore, as the heterogeneity level increases, the trade-off between fairness and accuracy degrades \cref{fig:fighetero_silo_var} and similarly, the trade-off worsens as the number of silos increases \cref{fig:fighetero_silo_var}. In the equalized odds experiments, \textit{SteFFLe} outperforms both \cite{ling2024fedfdp} and especially \cite{tran2021differentially} by such a wide margin that the scale of the plots becomes distorted, necessitating the exclusion of \cite{tran2021differentially} from the plots for clarity. Moreover while observing the cost of incorporating federated learning and differential privacy is observed through the gap beteween the four tradeoff curves. For $\varepsilon = 1$, \textit{FERMI} demonstrates the best trade-off, followed by \textit{non-private SteFFLe}, with \textit{DP-FERMI} and \textit{SteFFLe} trailing behind for both demographic parity and equalized odds objectives. As the privacy budget increases to $\varepsilon = 3$ and $\varepsilon = 9$, we observe that the private, non-federated methods begin to outperform their non-private, federated counterparts, with other trade-offs being in similar scenarios as before.

\subsubsection{Demographic Parity}
\label{ssec:demopair_app}

\begin{figure*}[h]
        \subfloat[$\varepsilon = 1$, \textit{Homogeneous} ($h = 0$)]{
            \includegraphics[width=.48\linewidth]{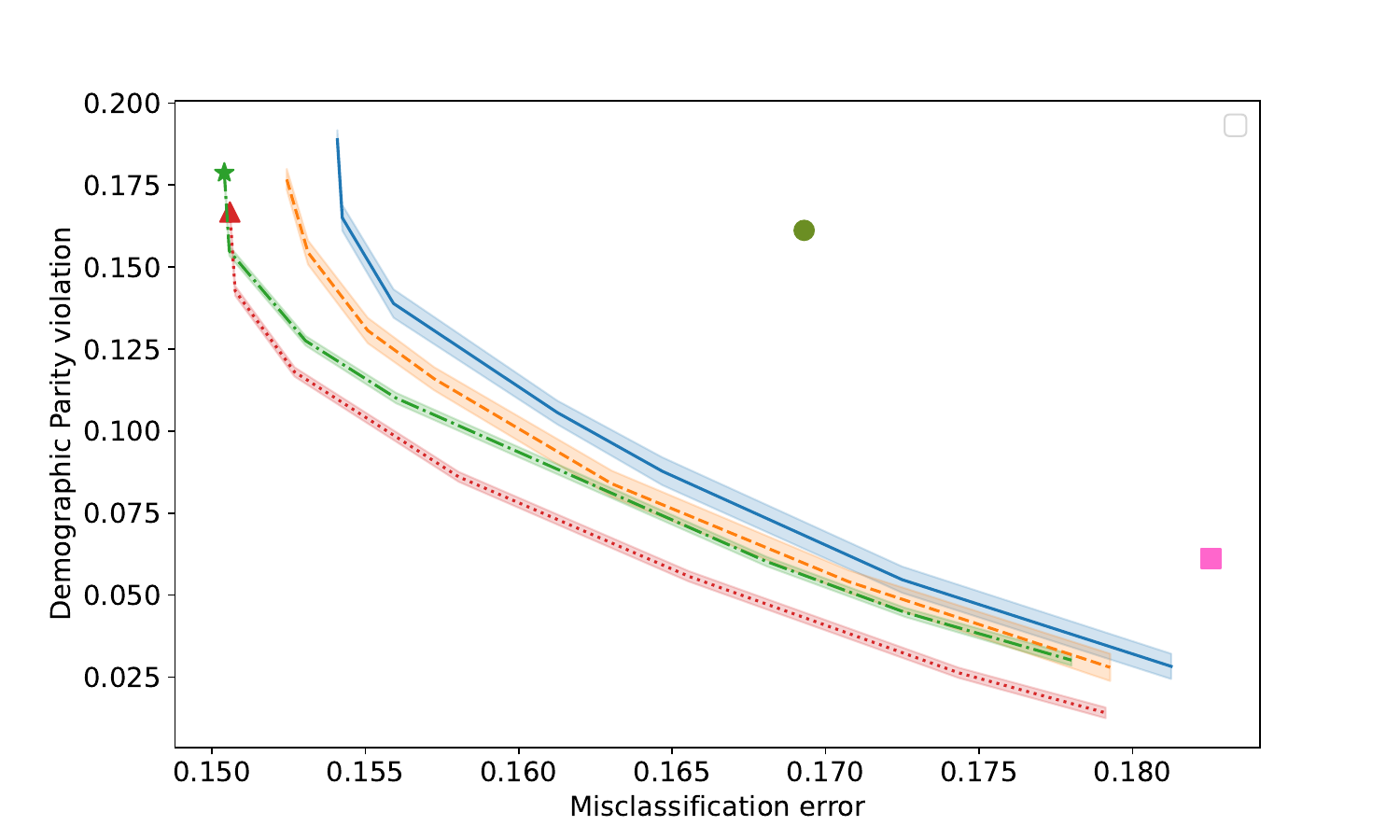}
            \label{subfig:a3}
        }\hfill
        \subfloat[$\varepsilon = 1$, \textit{Heterogeneous} ($h = 0.75$)]{
            \includegraphics[width=.48\linewidth]{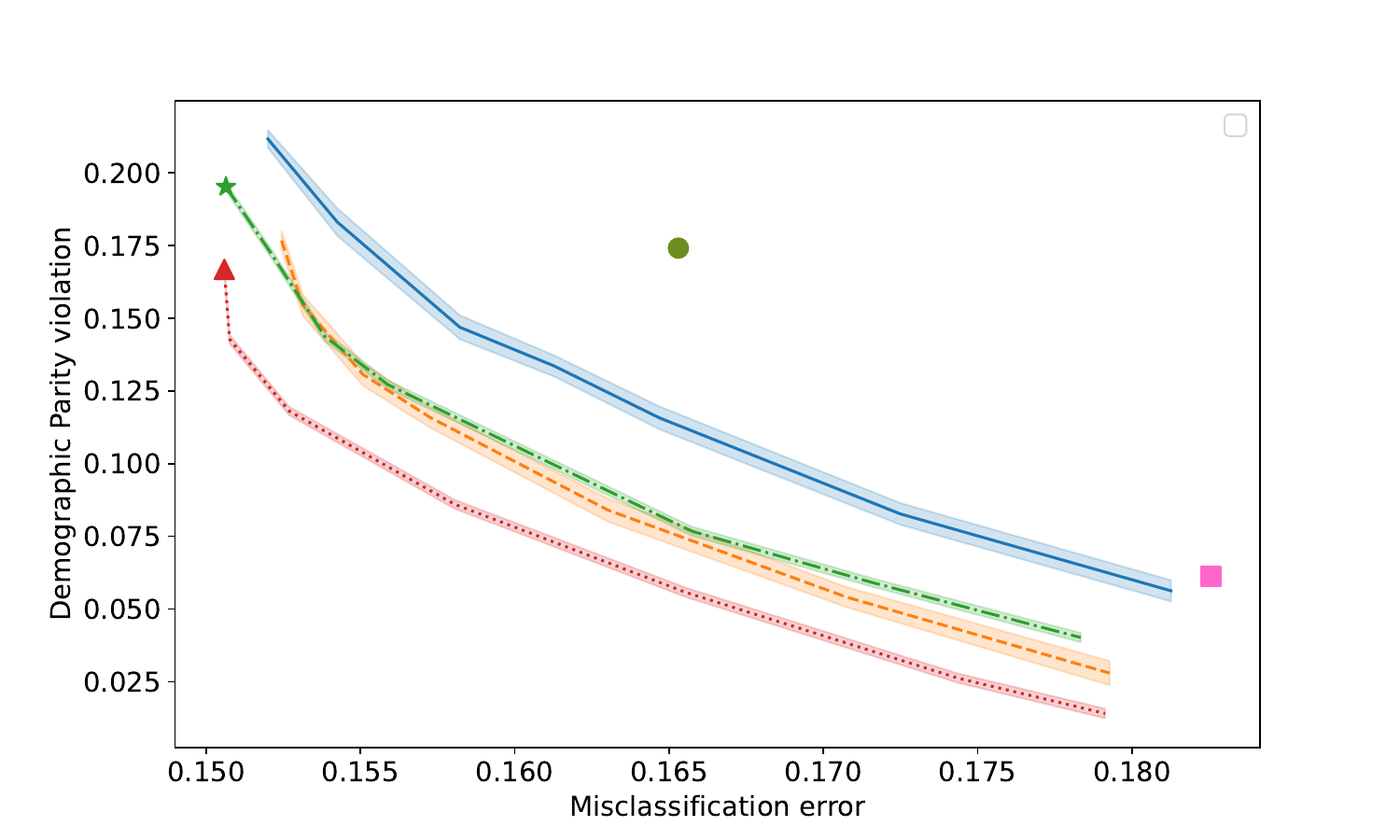}
            \label{subfig:b3}
        }\\
        \subfloat[$\varepsilon = 3$, \textit{Homogeneous} ($h = 0$)]{
            \includegraphics[width=.48\linewidth]{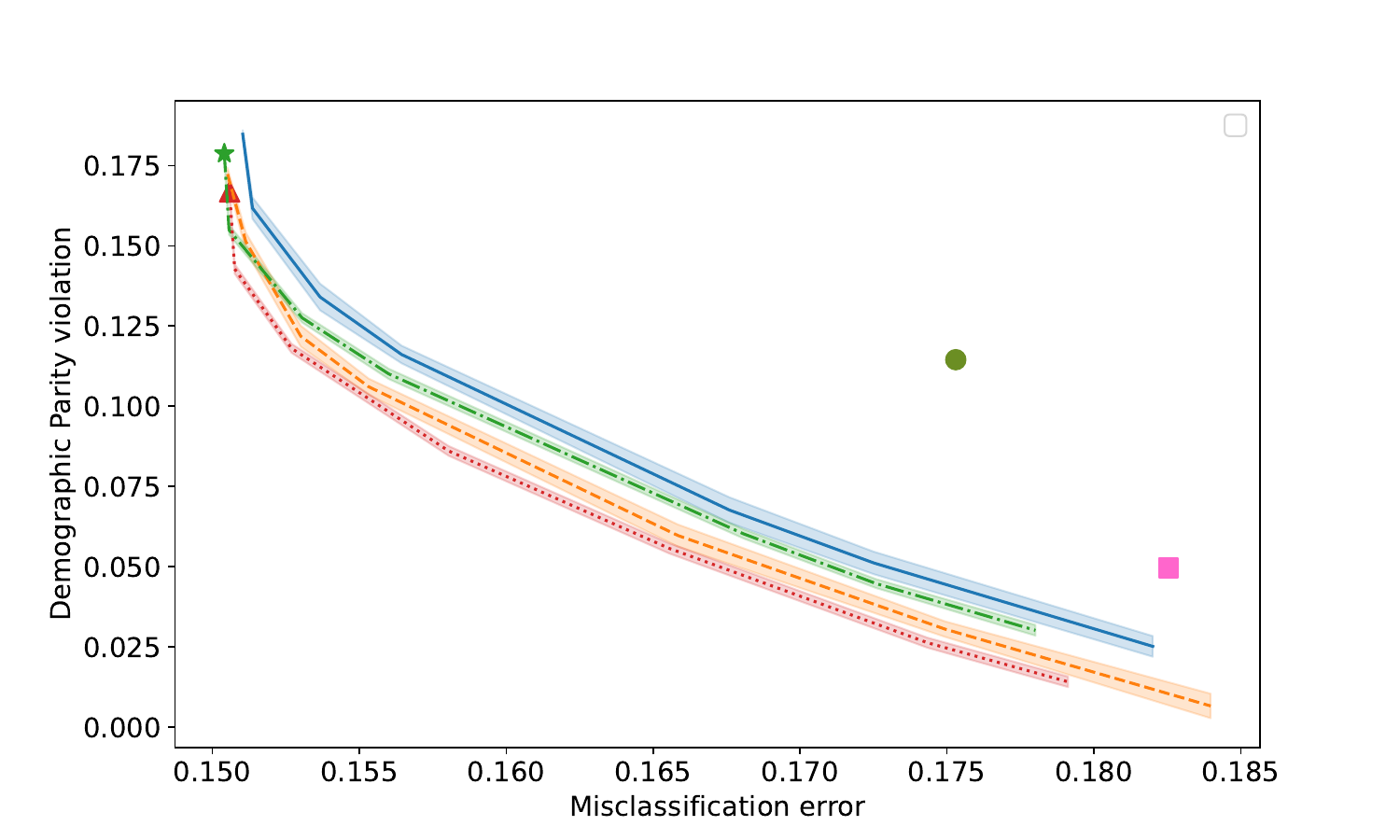}
            \label{subfig:c3}
        }\hfill
        \subfloat[$\varepsilon = 3$, \textit{Heterogeneous} ($h = 0.75$)]{
            \includegraphics[width=.48\linewidth]{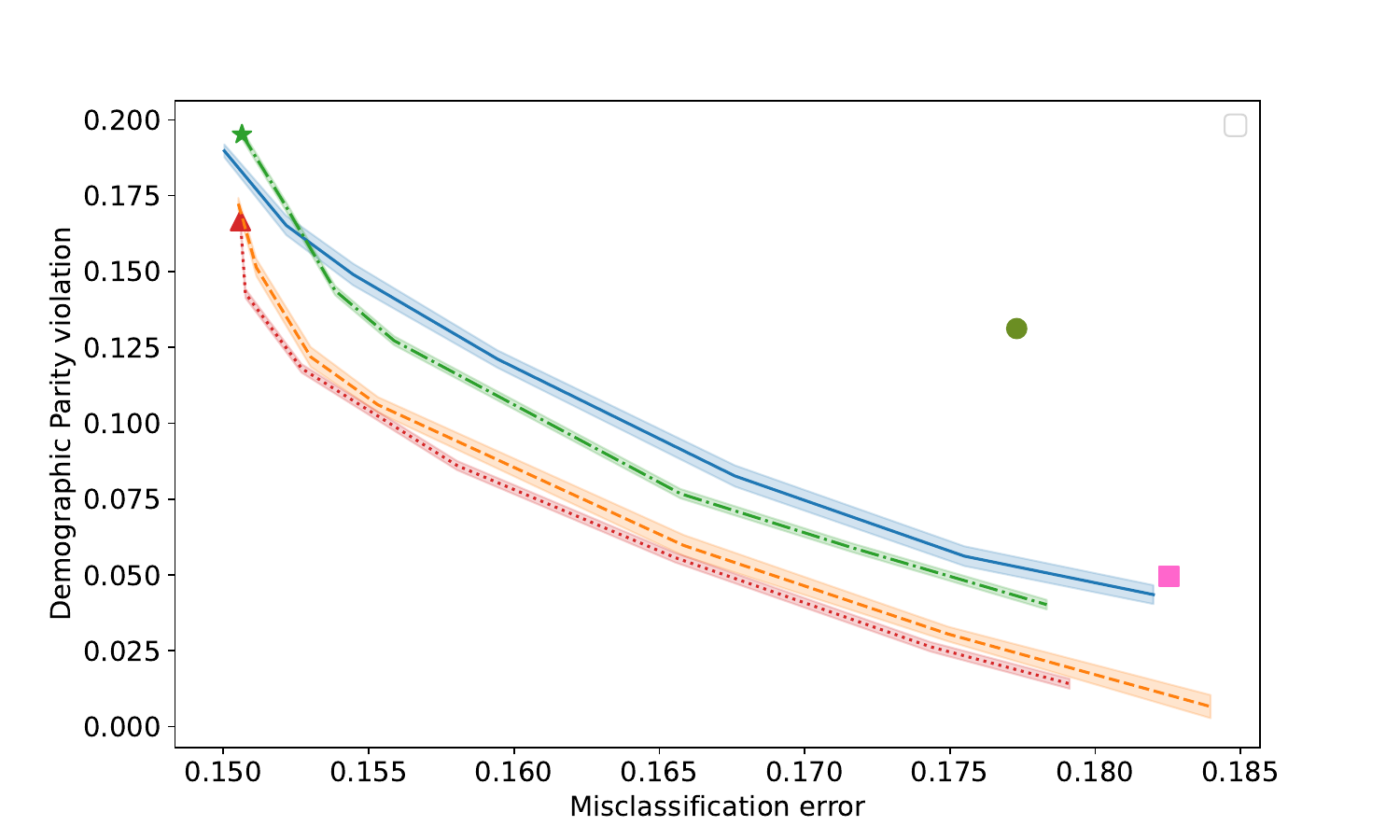}
            \label{subfig:d3}
        }
        \\
        \subfloat[$\varepsilon = 9$, \textit{Homogeneous} ($h = 0$)]{
            \includegraphics[width=.48\linewidth]{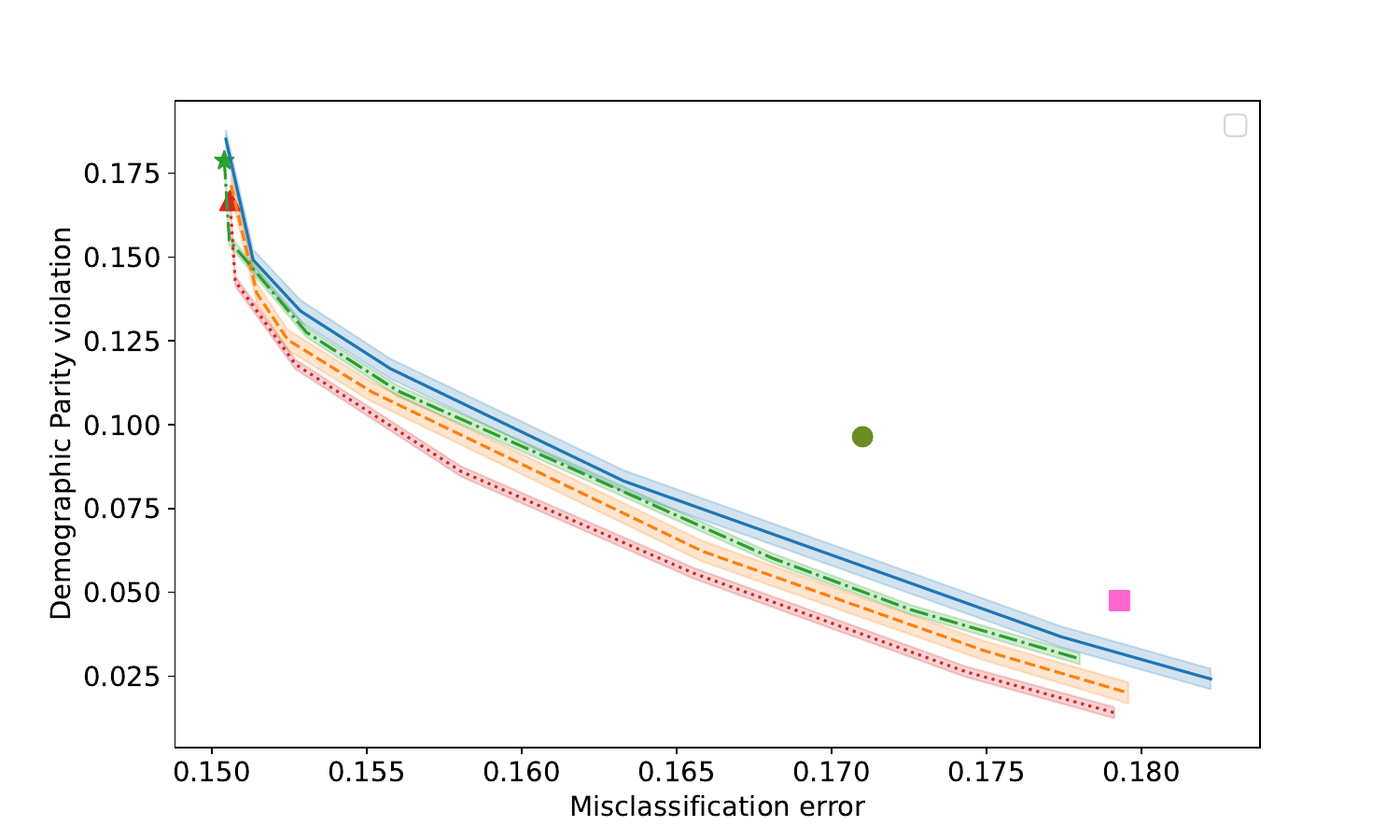}
            \label{subfig:e2}
        }\hfill
        \subfloat[$\varepsilon = 9$, \textit{Heterogeneous} ($h = 0.75$)]{
            \includegraphics[width=.48\linewidth]{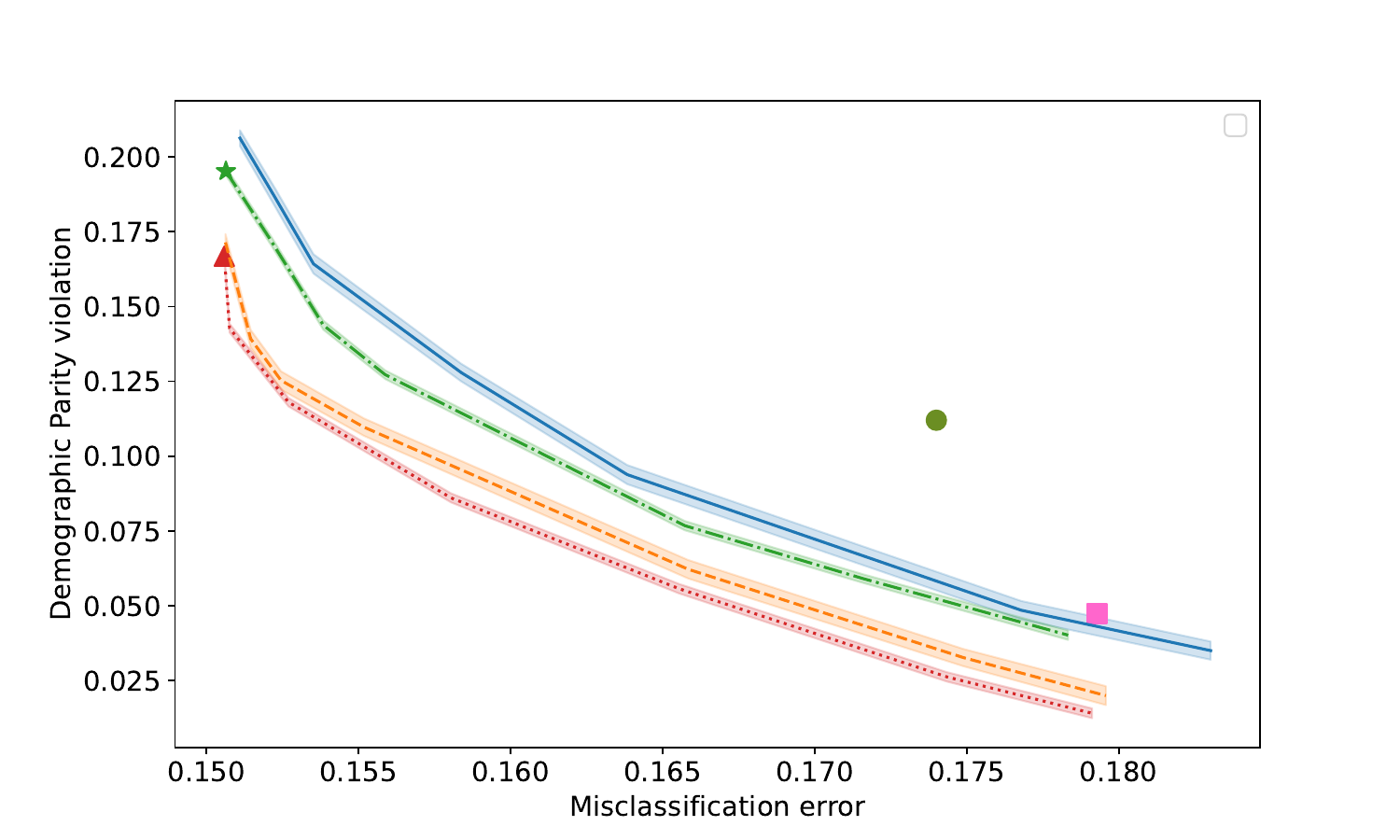}
            \label{subfig:f1}
        }\\
        \begin{center}
            \subfloat[Plot Legend]{
            \includegraphics[width=0.60\linewidth]{Experiments/final.pdf}%
            \label{subfig:g1}
        }
        \end{center}
        \caption{Demographic parity vs Misclassification error on \textit{Adult} dataset (\textit{Number of Silos} = 3)}
        \label{fig:figadultdp}
\end{figure*}

\FloatBarrier

\begin{figure*}[h]
        \subfloat[$\varepsilon = 1$, \textit{Homogeneous} ($h = 0$)]{
            \includegraphics[width=.48\linewidth]{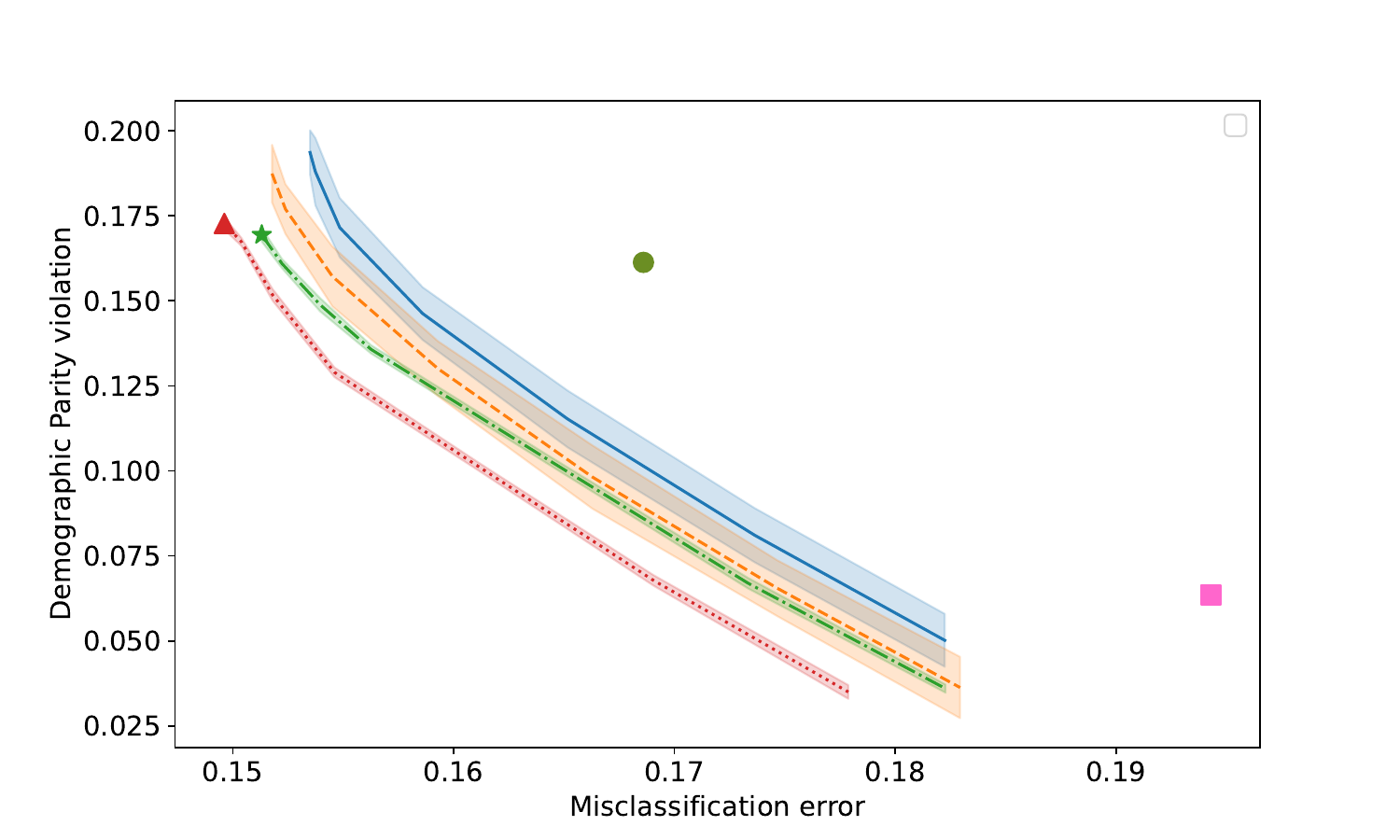}
            \label{subfig:a4}
        }\hfill
        \subfloat[$\varepsilon = 1$, \textit{Heterogeneous} ($h = 0.75$)]{
            \includegraphics[width=.48\linewidth]{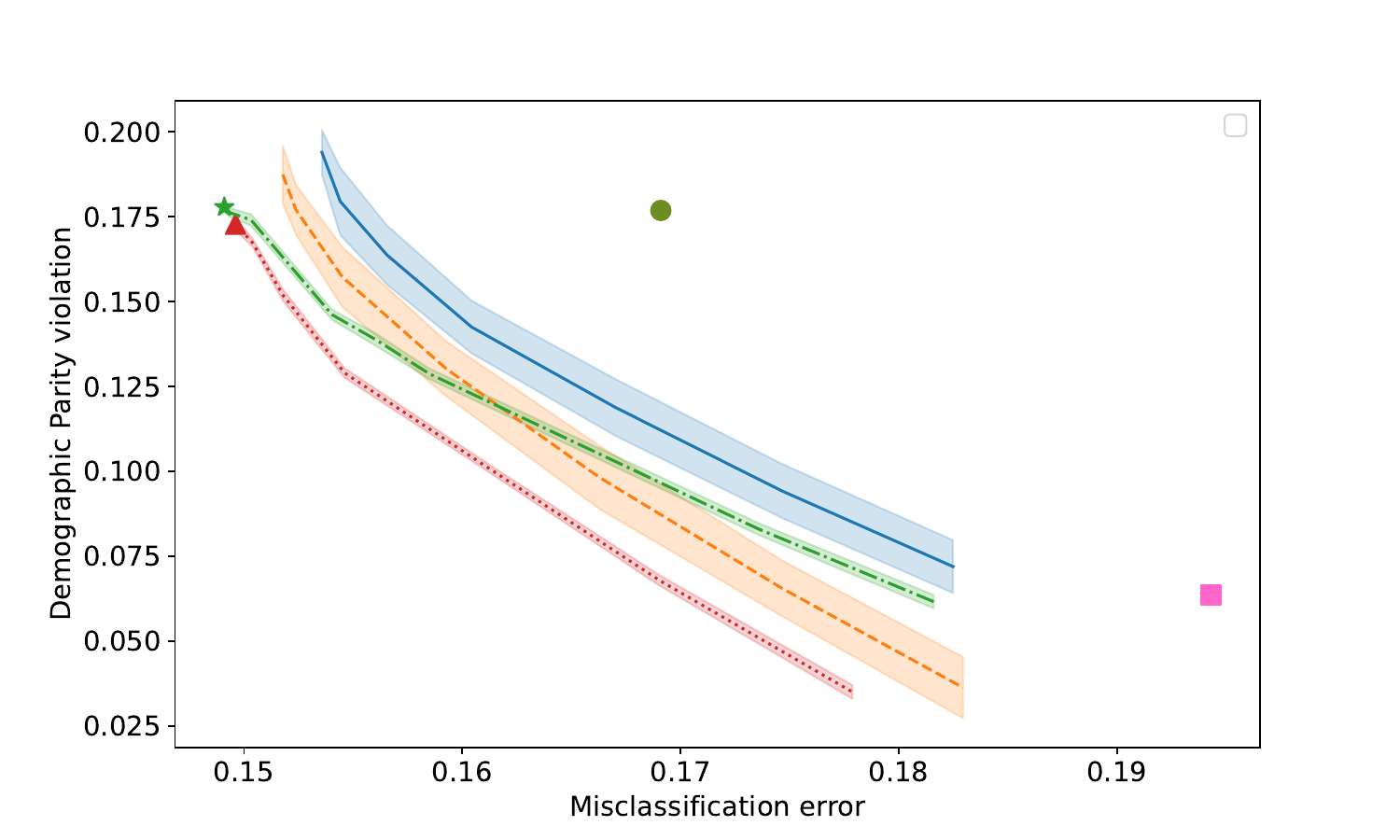}
            \label{subfig:b4}
        }\\
        \subfloat[$\varepsilon = 3$, \textit{Homogeneous} ($h = 0$)]{
            \includegraphics[width=.48\linewidth]{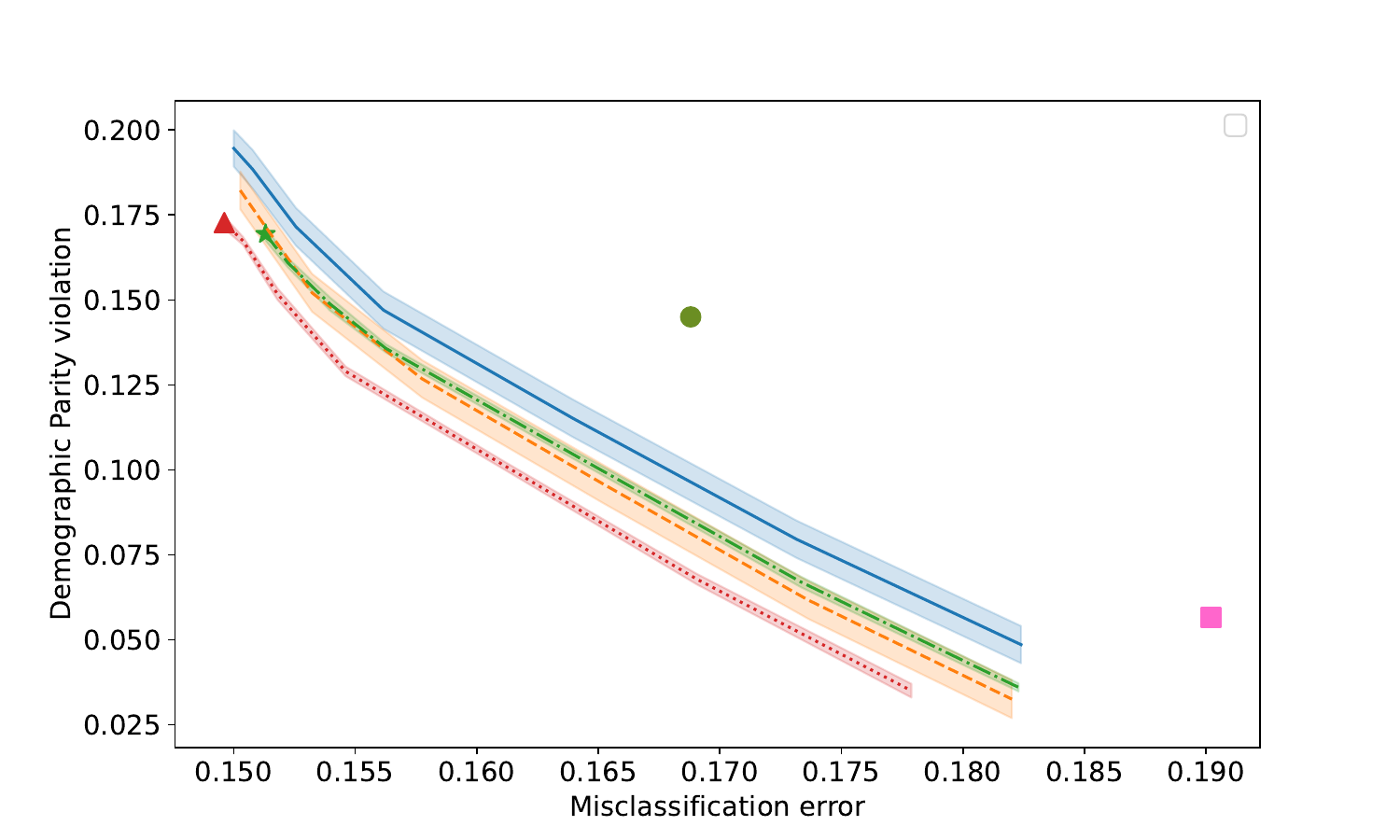}
            \label{subfig:c4}
        }\hfill
        \subfloat[$\varepsilon = 3$, \textit{Heterogeneous} ($h = 0.75$)]{
            \includegraphics[width=.48\linewidth]{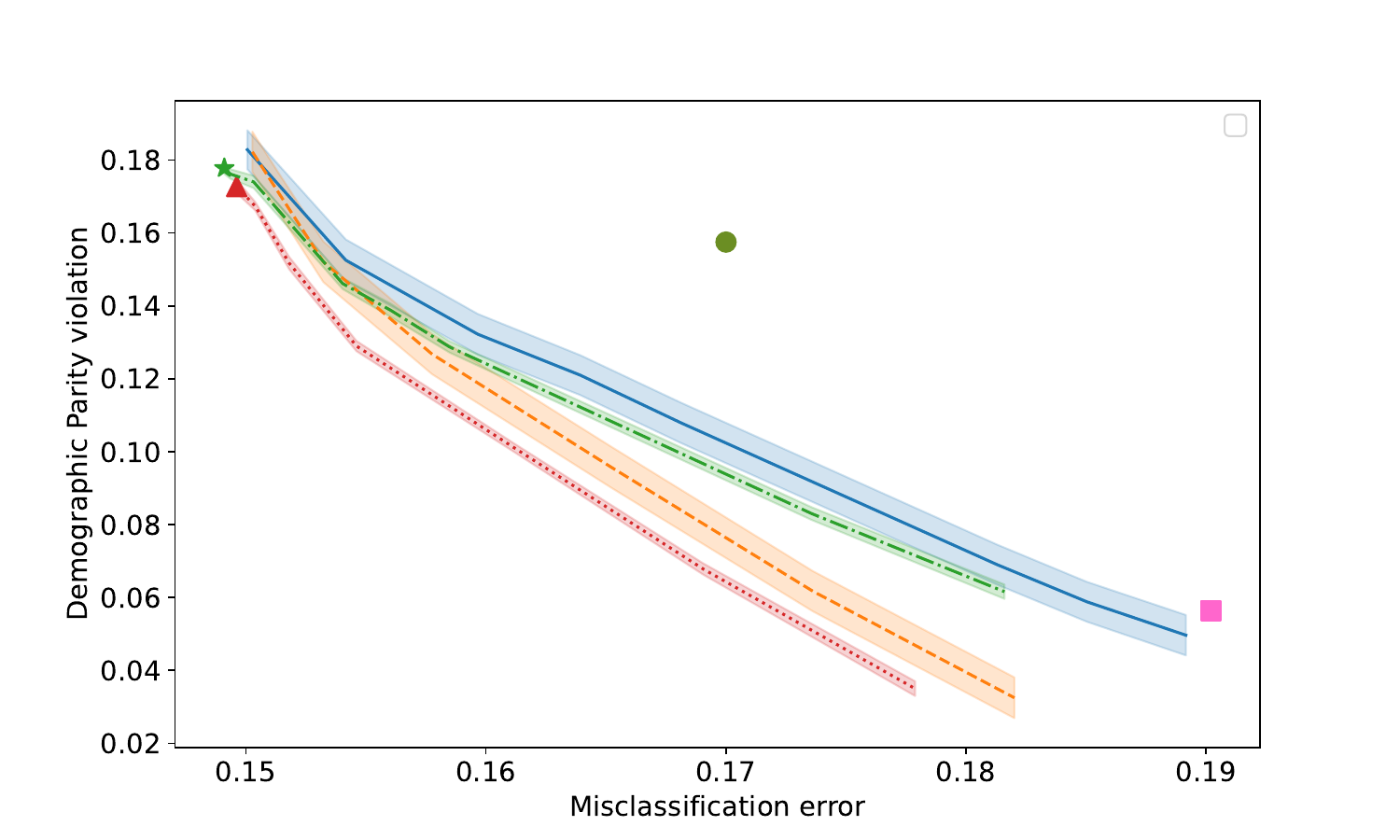}
            \label{subfig:d4}
        }
        \\
        \subfloat[$\varepsilon = 9$, \textit{Homogeneous} ($h = 0$)]{
            \includegraphics[width=.48\linewidth]{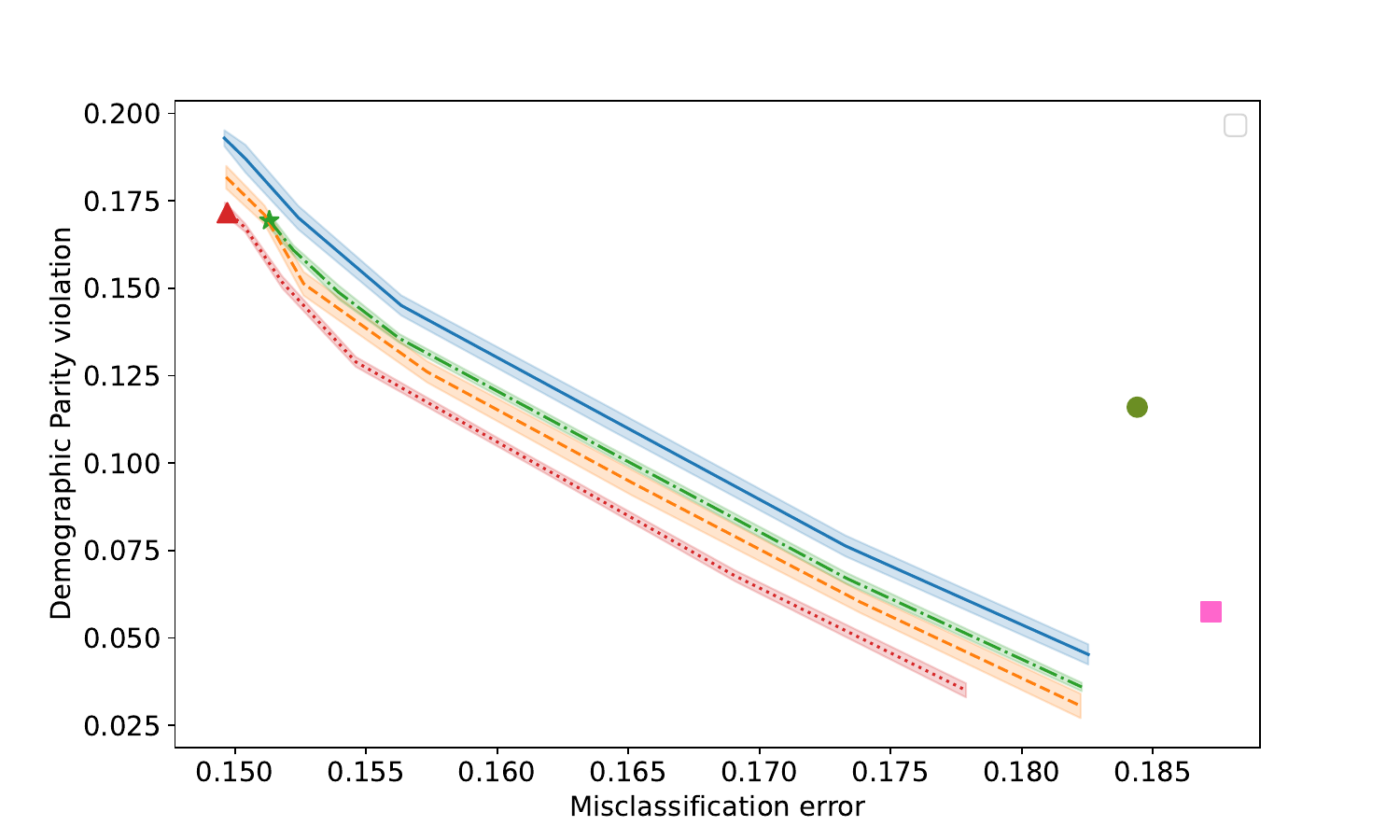}
            \label{subfig:e3}
        }\hfill
        \subfloat[$\varepsilon = 9$, \textit{Heterogeneous} ($h = 0.75$)]{
            \includegraphics[width=.48\linewidth]{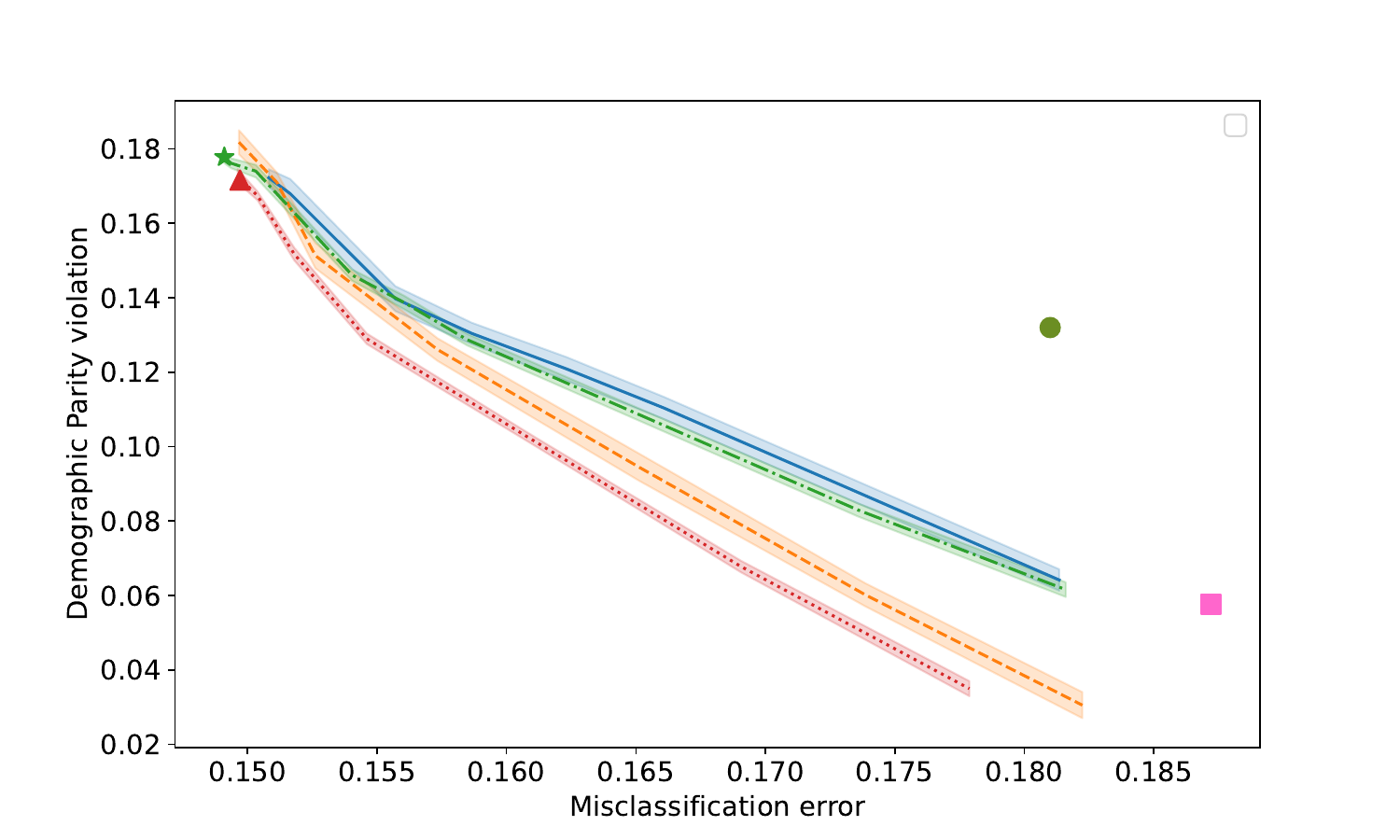}
            \label{subfig:f2}
        }\\
        \begin{center}
            \subfloat[Plot Legend]{
            \includegraphics[width=0.60\linewidth]{Experiments/final.pdf}%
            \label{subfig:g2}
        }
        \end{center}
        \caption{Demographic parity vs Misclassification error on \textit{Retired Adult} dataset (\textit{Number of Silos} = 3)}
        \label{fig:figretiredadultdp}
\end{figure*}

\FloatBarrier

\subsubsection{Equalized Odds}
\label{ssec:eo_app}

In this section, we now focus on a modified version of~\cref{alg: steffle}, which aims to minimize the violation of Equalized Odds. This is achieved by replacing the absolute probabilities in the objective function with class-conditional probabilities, as detailed in Equation \ref{eq:eqoddsviolation}.

For this set of experiments, we used the Credit Card dataset, keeping the sensitive attributes and output labels consistent with the previous sections. The results specific to the Credit Card dataset can be found in \cref{ssec:eo_app}. When compared to the methods proposed by~\citet{ling2024fedfdp} and the equalized odds objective introduced by~\citet{tran2021differentially}, \textit{our Equalized Odds variant of SteFFLe consistently outperforms these state-of-the-art baselines across all privacy and heterogeneity levels}. Notably, our model surpasses the performance of~\citet{tran2021differentially} by an exceptional margin—exceeding their results by over 150\%. This significant improvement resulted in a visual distortion of the corresponding plots, leading to their exclusion to preserve the clarity and interpretability of the visual representations.

\begin{figure*}[h]
        \subfloat[$\varepsilon = 1$, \textit{Homogeneous} ($h = 0$)]{
            \includegraphics[width=.48\linewidth]{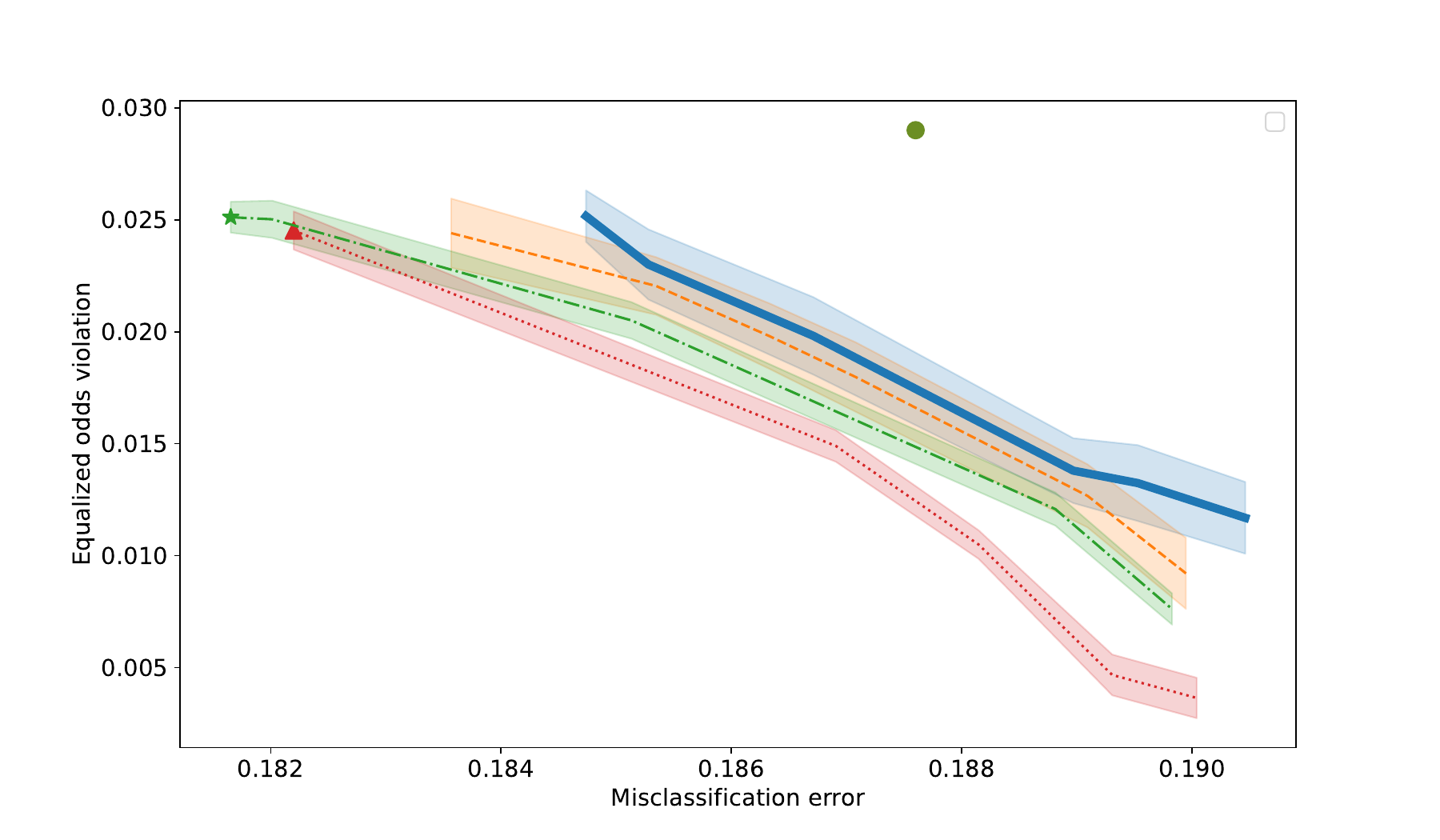}
            \label{subfig:a5}
        }\hfill
        \subfloat[$\varepsilon = 1$, \textit{Homogeneous} ($h = 0.75$)]{
            \includegraphics[width=.48\linewidth]{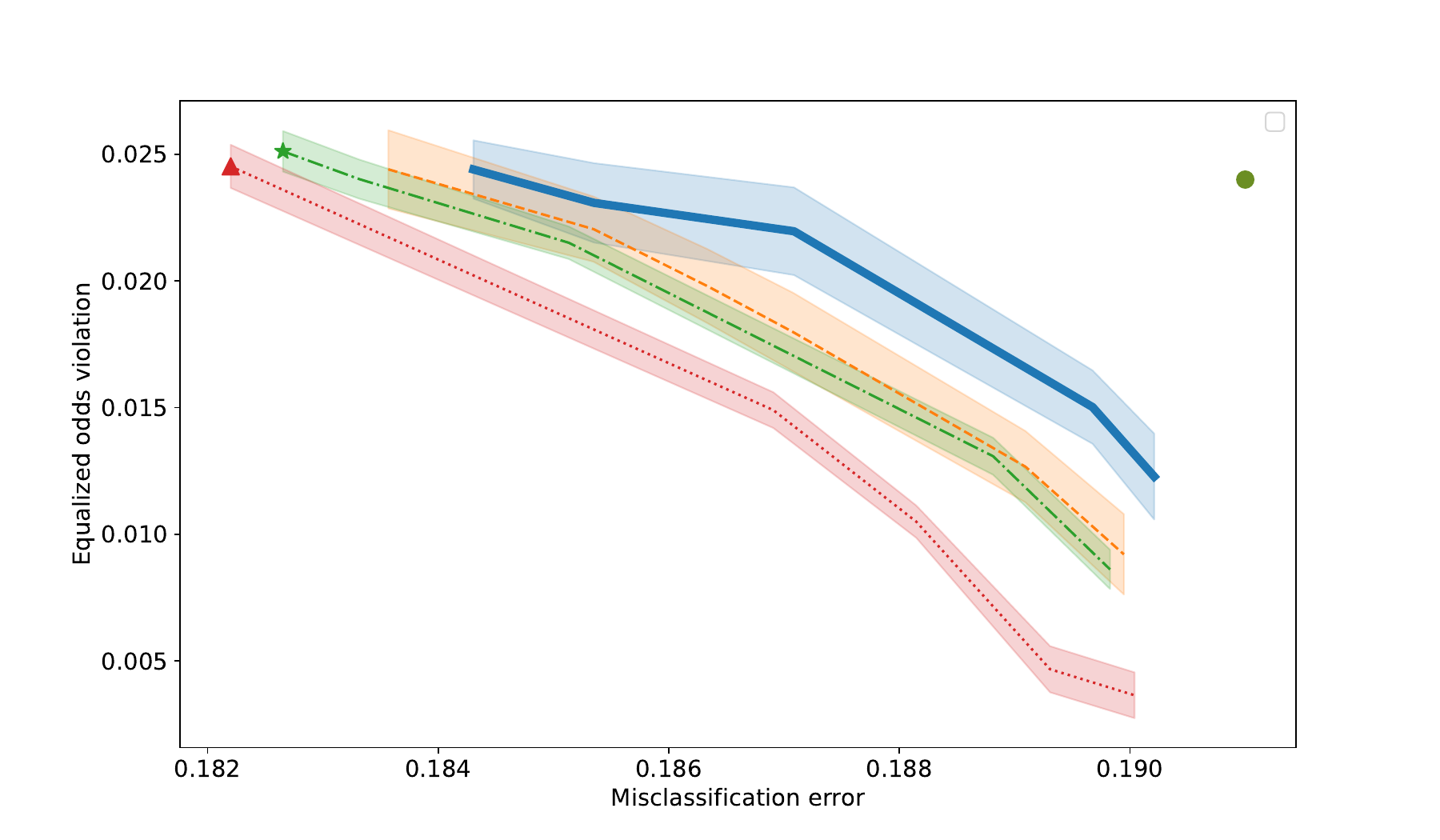}
            \label{subfig:b5}
        }\\
        \subfloat[$\varepsilon = 3$, \textit{Homogeneous} ($h = 0$)]{
            \includegraphics[width=.48\linewidth]{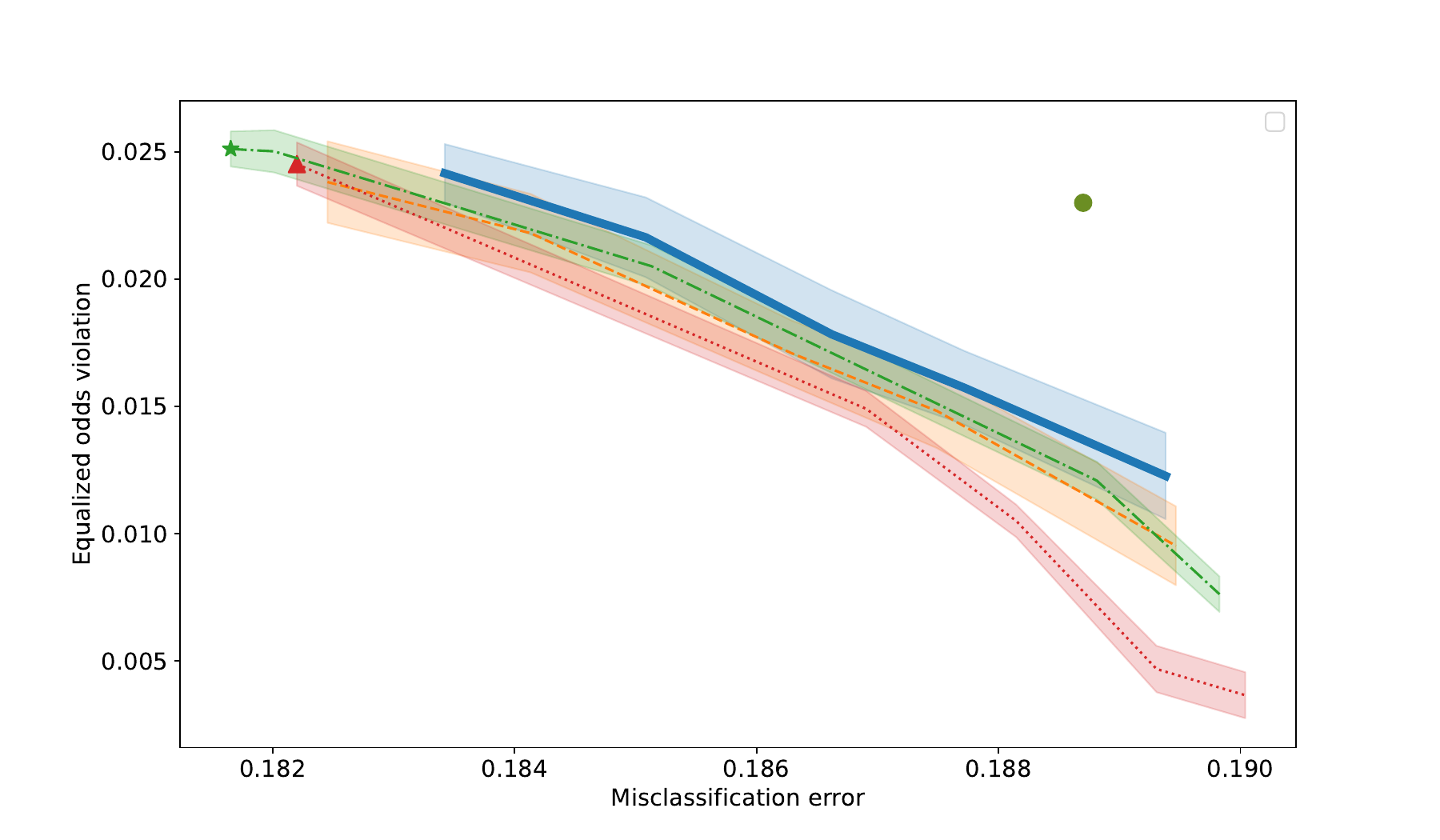}
            \label{subfig:c5}
        }\hfill
        \subfloat[$\varepsilon = 3$, \textit{Homogeneous} ($h = 0.75$)]{
            \includegraphics[width=.48\linewidth]{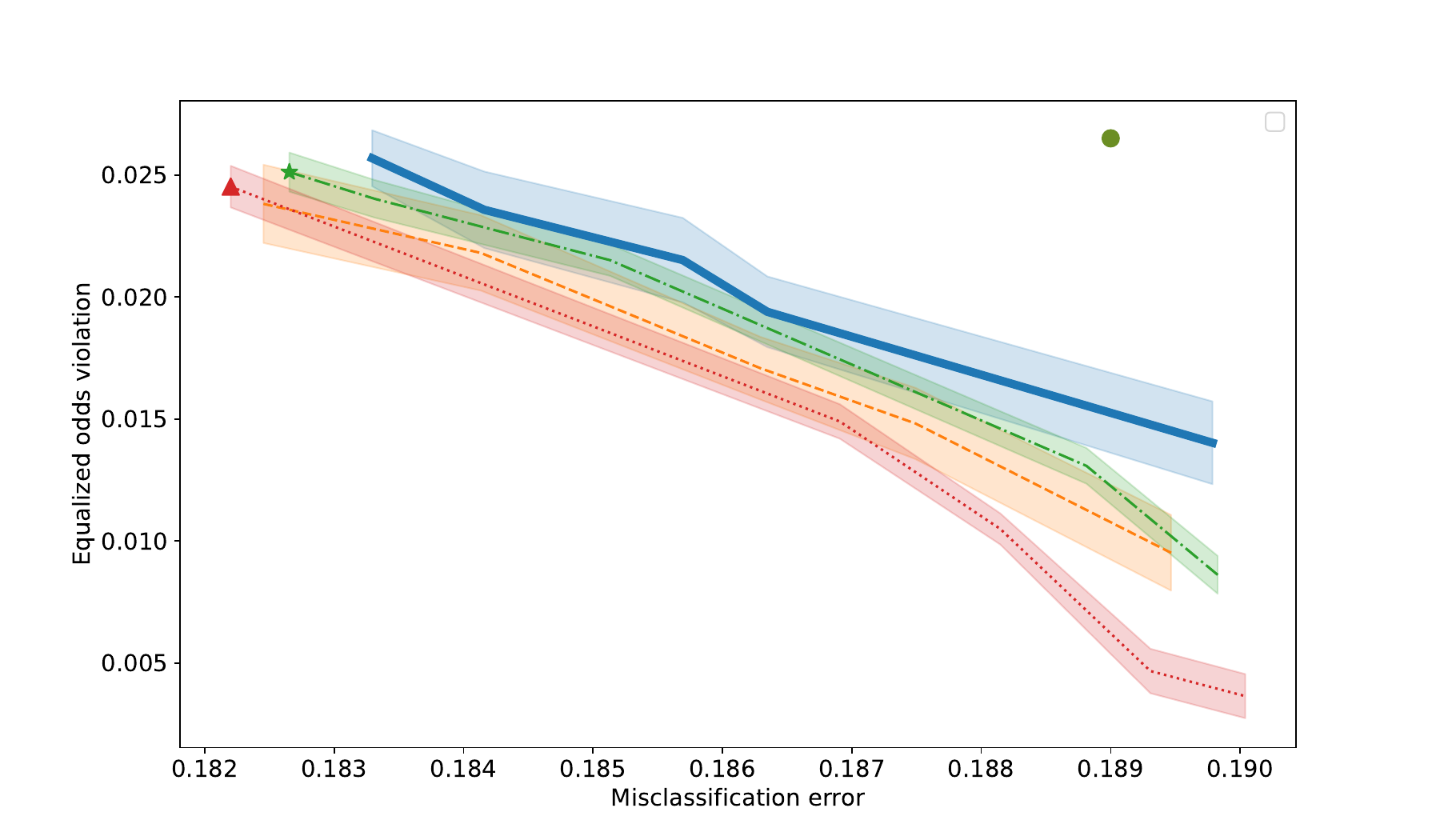}
            \label{subfig:d5}
        }
    `\begin{center}
            \subfloat[Plot Legend]{
            \includegraphics[width=0.60\linewidth]{Experiments/final.pdf}%
            \label{subfig:e4}
        }
        \end{center}
        \caption{Equalized Odds vs Misclassification error on \textit{Credit Card} dataset (\textit{Number of Silos} = 3)}
        \label{fig:figeqodds}
        
\end{figure*}

\FloatBarrier

\section{Additional Discussion and Examples of Different Modes of Centralization}
\label{examples:hybrid}

\paragraph{Centralized Sensitive Data and Decentralized Public Data.} An example of a silo containing centralized sensitive data is the \textit{United States Census Bureau}. It provides essential demographic, social, and economic data that various institutions utilize for a wide range of purposes. Some examples of these institutions include government agencies, academics, and non-profit organizations. The data with these institutions correspond to silos with public data and any machine learning model they train for decision making would require a combination of their own data and the centralized sensitive data provided by the Census.

\textbf{General Case: Arbitrary Numbers of Sensitive and Non-Sensitive Silos.} {Recall that every data-point we have is represented by a tuple $\{((x_u, y_u), s_u)\}_{u=1}^n$. We refer to $(x_u, y_u)$ as the non-sensitive part of the datapoint and $s_u$ to be the sensitive part. We assume that every silo indexes the data by universal index assigned to each data-point (say indexed by $1, 2, .., n$) instead of their local index. The data is distributed between the silos as follows: 
\begin{itemize}
    \item Let there be $p$ silos (represented by $1, ..., p$) with non-sensitive parts of the data (non-sensitive silos) and $s$ silos (represented by $1, ..., s$) with the sensitive parts of data (sensitive silos).
    \item Let $i \in [p]$ be the non-sensitive silo containing the non-sensitive attributes of datapoints which have indices $P_i \subset [n]$ such that $P_i \cap P_j = \phi$ for all $i,j \in [p]$ for $i \neq j$ and $\bigcup_{i=1}^p P_i = [n]$.
    \item Similarly, let any sensitive silo $i \in [s]$, contain the non-sensitive attributes of datapoints which have indices $S_i \subset [n]$ such that $S_i \cap S_j = \phi$ for all $i,j \in [s]$ for $i \neq j$ and $\bigcup_{i=1}^s S_i = [n]$.
\end{itemize}

For the training to happen, the non-sensitive silos locally sample a batch of data $\mathcal{J}_j \subset P_j$ for all $j \in [p]$. They compute the gradients of the loss with using their part of data and broadcast it to the server. For the gradient of the regularizer, each non-sensitive silo $j$ broadcasts $\mathcal{J}_j$ along with their respective model outputs. Then, the sensitive silos $c$ (such that $S_c \bigcap \bigcup_{j=1}^p \mathcal{J}_j \neq \emptyset$) which have the data corresponding to the indices sampled by all the non-sensitive silos ($S_c \bigcap \bigcup_{j=1}^p \mathcal{J}_j$) compute the gradient using the broadcasted outputs by the model and their local sensitive data. Then, these silos locally add noise according to batch size being $|S_c \bigcap \bigcup_{j=1}^p \mathcal{J}_j|$, and broadcast these noisy gradients to the server. The server aggregates both gradients from the non-sensitive and the sensitive silos and updates the model parameters.

It is important that for every sensitive silo $c$ scales its noise according to batch size being $|S_c \bigcap \bigcup_{j=1}^p \mathcal{J}_j|$ to preserve ISRL DP. However, since we have assumed that only the sensitive silos corresponding to the data will participate implying that $|S_c \bigcap \bigcup_{j=1}^p \mathcal{J}_j| \geq 1$. Hence, the upper bound on the stationarity gap would still exist with the value of batch size being one, thus still preserving a theoretical guarantee.}

However, this approach may suffer from a minor shortcoming: The attacker may figure out the silos which contain the data by looking at the silos that did not participate. We can curb that by prompting all the silos to broadcast a ``dummy" Gaussian noise with zero mean and finite variance. Since the noise has zero mean, the unbiasedness and finite variance of the inner gradient would still be preserved. However, such a technique can slow convergence because of the additional variance due to the dummy noise. It would be interesting to come up with a faster and a more consistent protocol that can deal with issue.

An example of this setting can be seen in hospital records. Hospitals maintain electronic health records (EHRs) that contain sensitive patient information, including medical history, diagnoses, treatments, and billing information. Due to the Health Insurance Portability and Accountability Act (HIPAA), access to these records is strictly controlled and typically limited to authorized personnel to protect patient privacy. Since an area can have many such hospitals, these hospitals act as sensitive decentralized silos of data and any institution (such as insurance agencies or bank loan providers) which requires information from such hospitals and since the respective institutions need such data to train their own models and the data that they posses act as the "public silo". We can also extend this application into settings where certain countries may not allow the sensitive data to leave the country (eg. E.U. GDPR) but may require a fair model for their decision driven processes.

\end{document}